\documentclass[journal, trans]{IEEEtran}
\usepackage{amsmath,epsfig}
\usepackage{algorithm}
\usepackage{algpseudocode}
\usepackage{booktabs}
\usepackage[numbers,sort&compress]{natbib}
\usepackage{amssymb}
\usepackage{color}
\usepackage{float}
\usepackage{adjustbox}
\ifCLASSINFOpdf
   \usepackage{graphicx}
\else
\fi
\usepackage{array}

  \usepackage[caption=false,font=normalsize,labelfont=sf,textfont=sf]{subfig}
\usepackage{url}

\begin{document}
%
\title{Locality and Structure Regularized Low Rank Representation for Hyperspectral Image Classification}
%
%

\author{Qi~Wang,~\IEEEmembership{Senior Member,~IEEE,}
        Xiang~He,
        and~Xuelong~Li,~\IEEEmembership{Fellow,~IEEE}
\thanks{This work was supported by the National Key R\&D Program of China under Grant 2017YFB1002202, National Natural Science Foundation of China under Grant 61773316, Natural Science Foundation of Shaanxi Province under Grant 2018KJXX-024, Fundamental Research Funds for the Central Universities under Grant 3102017AX010, and the Open Research Fund of Key Laboratory of Spectral Imaging Technology, Chinese Academy of Sciences.}
\thanks{Q. Wang is with the School of Computer Science, with the Center for Optical Imagery Analysis and Learning (OPTIMAL)
and with the Unmanned System Research Institute (USRI), Northwestern Polytechnical University,
Xi'an 710072, Shaanxi, China (e-mail: crabwq@gmail.com).}
\thanks{X. He is with the School of Computer Science and the Center for Optical Imagery Analysis and Learning, Northwestern Polytechnical University, Xi'an 710072, Shaanxi, China (e-mail: xianghe@mail.nwpu.edu.cn).}
\thanks{X. Li is with the Xi'an Institute of Optics and Precision Mechanics, Chinese Academy of Sciences, Xi'an 710119, Shaanxi, P. R. China and with the University of Chinese Academy of Sciences, Beijing 100049, P. R. China (e-mail: xuelong\_li@opt.ac.cn).
}

}

%
%

\markboth{IEEE TRANSACTIONS ON GEOSCIENCE AND REMOTE SENSING,~Vol.~57, No.~2, Feb~2019}%
{Shell \MakeLowercase{\textit{et al.}}: Bare Demo of IEEEtran.cls for IEEE Journals}
%



\maketitle

\begin{abstract}
Hyperspectral image (HSI) classification, which aims to assign an accurate label for hyperspectral pixels, has drawn great interest in recent years. Although low rank representation (LRR) has been used to classify HSI, its ability to segment each class from the whole HSI data has not been exploited fully yet. LRR has a good capacity to capture the underlying low-dimensional subspaces embedded in original data. However, there are still two drawbacks for LRR. First, LRR does not consider the local geometric structure within data, which makes the local correlation among neighboring data easily ignored. Second, the representation obtained by solving LRR is not discriminative enough to separate different data.  In this paper, a novel locality and structure regularized low rank representation (LSLRR) model is proposed for HSI classification. To overcome the above limitations, we present locality constraint criterion (LCC) and structure preserving strategy (SPS) to improve the classical LRR. Specifically, we introduce a new distance metric, which combines both spatial and spectral features, to explore the local similarity of pixels. Thus, the global and local structures of HSI data can be exploited sufficiently. Besides, we propose a structure constraint to make the representation have a near block-diagonal structure. This helps to determine the final classification labels directly. Extensive experiments have been conducted on three popular HSI datasets. And the experimental results demonstrate that the proposed LSLRR outperforms other state-of-the-art methods.
\end{abstract}


%
\IEEEpeerreviewmaketitle

\section{Introduction}
\label{sec:introduction}
%
%
%
%
\IEEEPARstart{H}{yperspectral} images (HSIs) are acquired by hyperspectral imaging sensors from the same spatial location and different spectral wavelengths. Due to the quite small wavelength interval (usually 10 $nm$) between every two neighboring bands, HSI generally has a very high spectral resolution. HSI is acquired from hundreds of continuous wavelengths, including a large range from visible to infrared spectrum, so HSI is composed of a great number of spectral bands, which makes hyperspectral data contain abundant discriminative information for the observed land surface. Since HSI can reflect well the distinct property of different land materials, HSI classification \cite{wang2017locality}, which is to assign the pixels of HSI a proper label, has attracted much attention over the past few decades.


Although the rich spectral information for each pixel brings a lot of help to classify hyperspectral data, there are still many challenges in HSI classification task. Due to the hundreds of spectral bands, the data of HSI has a very high dimensionality, which leads to the Hughes phenomenon \cite{hughes1968mean}. In addition, it usually costs lots of time to label HSI datasets, hence most hyperspectral data has very limited training samples, which becomes another major challenge. To address the above problems, a great number of SVM-based approaches have been developed over the past years. Support Vector Machine (SVM) is a widely used classifier in most classification tasks. Since it can effectively handle the high dimensional data, SVM has achieved great success in HSI classification. SVM with composite kernel (SVMCK) \cite{SVMCK_compare} was proposed to construct multiple composite kernels, which integrates both spectral and spatial information to enhance the classification performance. Specifically, the weighted kernels in \cite{SVMCK_compare} can effectively solve the problem that HSI usually has the limited labeled samples. Besides, SVM with graph kernel (SVMGK) \cite{camps2010spatio} developed a recursive graph kernel, which considered high-level spatial relationship rather than the simple pairwise relation. Besides the advantage that graph kernel is easy to compute, it can also be suitable for the small training data. However, SVM-based methods have a common drawback that their performance is easy to be influenced by parameters settings.

Motivated by recent development in subspace segmentation, low rank representation (LRR) has become an effective method for HSI classification.
LRR was first proposed for subspace segmentation by Liu \emph{et al}. in \cite{liu2010robust}.
Due to its considerable ability to exploit the underlying low-dimensional subspace structures of given data, LRR has attracted extensive attention and achieved great success in various fields, such as face recognition \cite{li2014learning}, image classification \cite{LRR_image}, subspace clustering \cite{liu2013robust}, object detection \cite{zhou2013moving}, etc. In particular, LRR is also applied successfully in hyperspectral image analysis \cite{wang2018getnet} and obtains promising performance in the past few years. For instance, Sun \emph{et al}. \cite{sun2014structured} presented a structured group low-rank prior, incorporating the spatial information, for sparse representation (SR) to classify HSI. Mei \emph{et al}. \cite{mei2016spectral} proposed to decompose the original hyperspectral data into low-rank intrinsic spectral signature and sparse noise to alleviate spectral variation, which degrades strongly the performance of hyperspectral analysis. However, there are still some shortcomings for common LRR. First, in spite that LRR has a great ability to capture the global structure of given data, it ignores the equally crucial local structure. This makes LRR fail to characterize the neighboring relation of each two pixels. Second, if all data are located in the union of multiple independent subspaces, the observed data with the same class should lie in the same subspace. Therefore, the ideal representation of given data would have a class-wise block-diagonal structure. Nevertheless, the traditional LRR can not obtain that structure. Third, most LRR based methods employ the whole samples as the dictionary to learn the low-rank representation. However, the dictionary has too many redundant atoms, which not only increases the computational cost, but also decreases the discriminative ability to reveal the potential property of HSI.

To tackle the aforementioned drawbacks, this paper proposes a novel locality and structure regularized low rank representation (LSLRR) for HSI classification. The main contributions are summarized as follows.

1) We introduce a new distance metric to measure the similarity of HSI pixels. For HSI classification, the spatial information is of great importance to acquire higher classification accuracy. The proposed measurement skillfully combines both spectral and spatial features into a unified distance metric, in which the involved parameter can be adjusted to fit different HSI datasets with different compactness of each class.

2) We present a novel locality constraint criterion (LCC) for LRR to further exploit the low-dimensional manifold structure of HSI. LRR can effectively capture the global structure of the given data, but the local geometry structure is also significant for most tasks. The proposed LSLRR with LCC successfully characterizes the global and local structures of HSI to explore the more reasonable representation.

3) An effective structure preserving strategy (SPS) is proposed to learn the more discriminative low-rank representation for HSI data. As we all know, the ideal representation of multi-class data has a class-wise block-diagonal structure. However, the original LRR hardly obtain the representation like that. Moreover, the learned representation for testing set can be used directly to classify HSI.

The reminder of this paper is organized as follows. In section \ref{sec:related_work}, two typical representation-based methods for hyperspectral image analysis are introduced. Then the proposed LSLRR is described in detail in section \ref{sec:LSLRR}. An optimization algorithm for solving LSLRR is derived in section \ref{sec:solution}. Besides, section \ref{sec:experiments} shows the extensive experimental results and corresponding analyses. Finally, we conclude this paper in section \ref{sec:conclusion}.

\section{Related Work}
\label{sec:related_work}
As we all know, low rank representation (LRR) and sparse representation (SR) are two typical representation-based approaches. This paper mainly focuses on LRR, which has achieved huge success in hyperspectral remote sensing fields \cite{wang2016salient}. Since SR has some common features with LRR, and has also attracted much attention in recent years, we will provide an overview about both SR-based and LRR-based methods for HSI classification in this section. In addition, the involved dictionary learning techniques are also introduced here.

\subsection{SR-based Methods}
Given some data vectors, SR seeks the sparse representation based on the linear combination of atoms in dictionary. Due to its great classification performance, SR has been applied widely in hyperspectral analysis. Chen \emph{et al}. \cite{chen2011hyperspectral} proposed a joint sparsity model which represented the hyperspectral pixels within a patch by the same sparse coefficients. In \cite{srinivas2013exploiting}, the sparsity of HSI was exploited by a probabilistic graphical model, which can effectively capture the conditional dependences. Zhang \emph{et al}. \cite{zhang2014nonlocal} developed a nonlocal weighted joint SR model, where different weights were employed to spatial neighboring pixels. In order to solve the problem that SR-based methods usually neglect the representation residuals, Li \emph{et al}. \cite{li2016hyperspectral} proposed a robust sparse representation for HSI classification, which is robust for outliers. Moreover, Li \emph{et al}. \cite{li2015efficient} presented a new superpixel-level joint sparse model (JSM) for HSI classification, which explored the class-level sparsity to combine multiple-features of pixels in local regions. A spectral-spatial adaptive SR was developed for HSI compression in \cite{fu2017adaptive}, which made use of both spectral and spatial features. And it utilized superpixel segmentation to generate adaptive homogeneous regions. Gan \emph{et al}. \cite{gan2018multiple} incorporated multiple types of features, which helps so much for HSI classification task, into a kernel sparse representation classifier (KSRC). In addition, Fang \emph{et al}. \cite{fang2014spectral} proposed a multiscale adaptive sparse representation, which effectively integrated contextual feature at multiple scales by an adaptive sparse technique. Considering that $\ell_1$-based SR may obtain unstable representation results, Tang \emph{et al}. \cite{tang2016manifold} incorporated manifold learning into SR to exploit the local structure and get the smooth sample representation. For more detailed description, A useful survey about SR-based methods can be referred in \cite{li2016survey}.

\subsection{LRR-based Methods}
Another popular representation-based method is LRR. Different from SR, LRR seeks the low-rank representation for given data. And most LRR-based approaches have been proposed for hyperspectral image analysis \cite{wang2018optimal}. Du \emph{et al}. \cite{qu2014abundance} utilized the joint sparse and low rank representation to solve the abundance estimation problem for HSI. Low-rank constraint is integrated to overcome the drawback of local spectral redundancy and correlation for HSI denoising in \cite{zhao2015hyperspectral}. Shi \emph{et al}. \cite{shi2015domain} proposed a semi-supervised framework for HSI classification, where LRR reconstruction is employed to decrease the influence of noise and outliers and make domain adaption more robust. A novel framework combining the maximum a posteriori (MAP) and LRR, exploiting the high spectral correlation, is proposed for HSI segmentation in \cite{yuan2015low}. Considering that the underlying low-dimensional structure in HSI data is multiple subspaces rather other single subspace, Sumarsono \emph{et al}. \cite{sumarsono2016low} adopted LRR as a preprocessing step for supervised and unsupervised classification of HSI. Most studies have demonstrated that the contextual information is very beneficial to improve the classification accuracy of HSI. Almost all state-of-the-art work, which employed LRR for HSI classification, combined both spectral and spatial features. For instance, a new low-rank structured group priori was presented to exploit the spatial information between neighboring pixels by Sun \emph{et al}. in \cite{sun2014structured}. Soltani-Farani \emph{et al}. \cite{soltani2015spatial} proposed to add the spatial characteristics by partitioning the HSI into several square patches as contextual groups. However, the fixed-size squares window neglects the difference between the pixels in the same window. He \emph{et al}. \cite{he2016low} applied a superpixel segmentation algorithm to divide HSI into some homogeneous regions with adaptive size, which is better than fixed-size patches to utilize contextual features. In addition, a new spectral-spatial HSI classification method using $\ell_{1/2}$ regularized LRR was developed in \cite{jia2015spectral}, where the contextual information is efficiently incorporated into the spectral signatures by representing the spatial adjacent pixels in a low-rank form.

\subsection{Dictionary Learning}
Since LRR can greatly exploit the global structure for the given data, it is superior to SR in some cases. Even so, one thing that LRR and SR have in common is that they both assume to describe every sample as the linear combination of some atoms in a given dictionary. And the selection of dictionary is fairly important to the performance of LRR. In general, dictionary learning methods can be roughly divided into two categories \cite{rubinstein2010dictionaries}: (1) learning a dictionary based on mathematical model. Many traditional models such as contourlet, wavelet, bandelet, wavelet packets, all can be used to construct an effective dictionary. (2) building a dictionary to behave well in training set. The second class of methods have brought more and more concern. The major advantage is that they can obtain great experimental results in most practical applications. These state-of-the-art methods include Optimal Directions (MOD) \cite{engan1999method}, Union of Orthobases \cite{lesage2005learning}, Generalized PCA (GPCA) \cite{vidal2005generalized}, K-SVD \cite{aharon2006rm} and so on. For HSI classification, some dictionary learning techniques have been proposed. Soltani-Farani \emph{et al}. \cite{soltani2015spatial} presented a spatial-aware dictionary learning method that is to divide HSI data into some contextual neighborhoods and then model the pixels with the same group as a common subspace. Motivated by Learning Vector Quantization (LVQ), Wang \emph{et al}. \cite{wang2014spatial} proposed a novel dictionary learning method for the sparse representation, and modeled the spatial context by a Bayesian graph. He \emph{et al}. \cite{he2016low} applied a joint low rank representation model in every spatial group to learn an appropriate dictionary.

\section{Locality and Structure Regularized Low Rank Representation (LSLRR)}
\label{sec:LSLRR}
In this section, we will describe the proposed LSLRR in detail. The original LRR formulas are first introduced. Then two main powerful regularization terms and dictionary learning scheme are presented. Finally, we derive an optimization algorithm to solve the objective function of LSLRR.

\subsection{Low Rank Representation}
Low rank representation (LRR) is based on the assumption that all data are sufficiently sampled from multiple low-dimensional subspaces embedded in a high-dimensional space. \cite{liu2010robust} indicates that LRR can effectively explore the underlying low-dimensional structures for the given data. Assume that data samples $Y \in \mathbb{R}^{d \times n}$ are drawn from a union of many subspaces which are denoted as $\bigcup^k_{i=1}S_k$, where $S_1, S_2,...,S_k$ are the low-dimensional subspaces. The LRR model aims to seek the low-rank representation $Z \in \mathbb{R}^{m \times n}$ and the sparse noises $E \in \mathbb{R}^{d \times n}$ based on the given dictionary $A \in \mathbb{R}^{d \times m}$. Specifically, LRR is formulated as the following rank minimization problem
\begin{align}
\label{Eq:LRR 0 norm}
\mathop{\min}\limits_{Z,E} \quad rank(Z)+\lambda \lVert E \rVert_0  \quad {s.t.} \quad  Y=AZ+E,
\end{align}
where $A$ and $E$ are the dictionary matrix and sparse noise component, respectively. $\lVert \cdot \rVert _0$ is the $\ell_0$ norm, the number of all nonzero elements. $\lambda$ is the regularization coefficient to balance the weights of rank term and reconstruction error. It is worth noting that the only difference between SR and LRR is that SR aims to find the sparsest representation while LRR is to seek the low-rank representation. But LRR can effectively capture the global structure of data samples.

However, it is difficult to solve the non-convex problem (\ref{Eq:LRR 0 norm}) due to the discrete nature of the rank operation and $\ell_0$ norm. Therefore, the original minimization problem (\ref{Eq:LRR 0 norm}) needs to be relaxed in order to make it solvable. The common convex relaxation of problem (\ref{Eq:LRR 0 norm}) is presented as
\begin{align}
\label{Eq:LRR 1 norm}
\mathop{\min}\limits_{Z,E} \quad \lVert Z \rVert_*+\lambda \lVert E \rVert_1  \quad {s.t.} \quad  Y=AZ+E,
\end{align}
where $\lVert \cdot \rVert _*$, defined as the sum of all singular values of $Z$, is the nuclear norm. $\lVert\cdot\rVert_1$ is the $\ell_1$ norm, i.e., the sum of the absolute value of all elements. And $\lVert Z \rVert_*$ and $\lVert E \rVert_1$ are the convex envelope of $rank(Z)$ and $\lVert E \rVert_0$, respectively. Then problem (\ref{Eq:LRR 1 norm}) has a nontrivial solution. In fact, the solution of problem (\ref{Eq:LRR 1 norm}) is equal to that of problem (\ref{Eq:LRR 0 norm}) in this case of free noise \cite{liu2013robust}. However, in practical applications most data are noisy, even strongly corrupted. Therefore, when a large number of data samples are grossly corrupted, a robust model \cite{liu2010robust} is presented as
\begin{align}
\label{Eq:LRR 2,1 norm}
\mathop{\min}\limits_{Z,E} \quad \lVert Z \rVert_*+\lambda \lVert E \rVert_{2,1}  \quad {s.t.} \quad  Y=AZ+E,
\end{align}
where $\lVert \cdot \rVert_{2,1}$ is the $\ell_{2,1}$ norm, which is defined as $\lVert E \rVert_{2,1} =\sum_{j}^{n}{\sqrt{\sum_{i}^{d} E_{i,j}^2}}$. Specifically, compared to $\ell_1$ norm, $\ell_{2,1}$ norm expects more columns of $E$ to be zero vector, i.e., some samples are clean and others are noisy.

\subsection{Locality Constraint Criterion (LCC) for LSLRR}
For hyperspectral image (HSI) classification, if some pixels have a neighboring relation, there is a high probability that they belong to the same class. That is, spatial similarity is a beneficial information to improve the classification accuracy of HSI. Therefore, it is very necessary to incorporate the contextual information into the classifier. Furthermore, LRR has a powerful ability to exploit the global structure of HSI data, but the local manifold structure between adjacent pixels, which is also helpful to classify HSI, is neglected by LRR. Therefore, we develop a local structure constraint, which utilizes both the spectral and spatial similarity, to improve the performance of the original LRR model.

Suppose that HSI data is denoted as $X=[x_1,x_2,...,x_n]\in \mathbb{R}^{d \times n}$, where $d$ and $n$ are the number of spectral bands and all pixels, respectively. And $x_i$ denotes the spectral column vector of the \emph{i}-th pixel of HSI data $X$. Similarly, assume that the spatial feature matrix $L=[l_1,l_2,...,l_n]\in \mathbb{R}^{2 \times n}$, and $l_i$ denotes the position coordinate of the \emph{i}-th pixel. A simple way to compute the distance matrix which combines both spectral and spatial features is formulated as
\begin{align}
\label{Eq:distance 1}
M_{ij}=\sqrt{\lVert x_i-x_j \rVert^2_2+\lVert l_i-l_j \rVert^2_2},
\end{align}
where $M_{ij}$ is the distance between the \emph{i}-th and \emph{j}-th pixels. Note that the spectral values of $X$ and coordinate values of $L$ are normalized to a range of [0, 1]. However, the distance metric is not reasonable enough because the above spectral and spatial features are unequal and have different physical meanings. Therefore, a more accurate similarity metric between two pixels is proposed as
\begin{align}
\label{Eq:distance 2}
M_{ij}=\sqrt{\lVert x_i-x_j \rVert^2_2+m\lVert l_i-l_j \rVert^2_2},
\end{align}
where $m$ is a hyper-parameter for controlling the weight of spectral and spatial distance. For different HSI datasets, the compactness of each category is different. And it is more appropriate to choose a large value of $m$ for the HSI dataset with high compactness of each class. As we all know, two pixels with a larger distance should have a smaller similarity. Besides, the low-rank representation $Z$ can be viewed as the affinity matrix, in which $Z_{ij}$ denotes the similarity of the \emph{i}-th and \emph{j}-th samples. As such, to keep the difference between classes and the compactness within classes, the locality constraint as a penalty term for LRR is introduced as follows
\begin{align}
\label{Eq:locailty constraint}
\sum_{i,j}{M_{ij}|Z_{ij}|}=\lVert M\circ Z\rVert_1,
\end{align}
where $\circ$ is the Hadamard product which denotes element-wise product of two matrixs. Moreover, the locality constraint also takes the sparsity of low-rank representation matrix $Z$ into account. Because $Z$ stands for the similarity between dictionary and the original data, all elements of $Z$ should have non-negative values. Therefore, the final locality regularization term can be written as $\lVert M\circ Z\rVert_1$ with the constraint $Z\geq 0$. And locality regularized low rank representation (LLRR) model can be formulated as
\begin{equation}
	\begin{split}
	\label{Eq:LLRR}
	&\mathop{\min}\limits_{Z,E} \quad \lVert Z \rVert_*+\lambda \lVert E \rVert_{2,1}+\alpha\lVert M\circ Z\rVert_1 \\
	&\text{  } s.t. \quad  Y=AZ+E,  Z\geq 0.
	\end{split}
\end{equation}

\begin{figure*}[htb]
	\begin{minipage}[b]{1.0\linewidth}
		\centering
		\centerline{\epsfig{figure=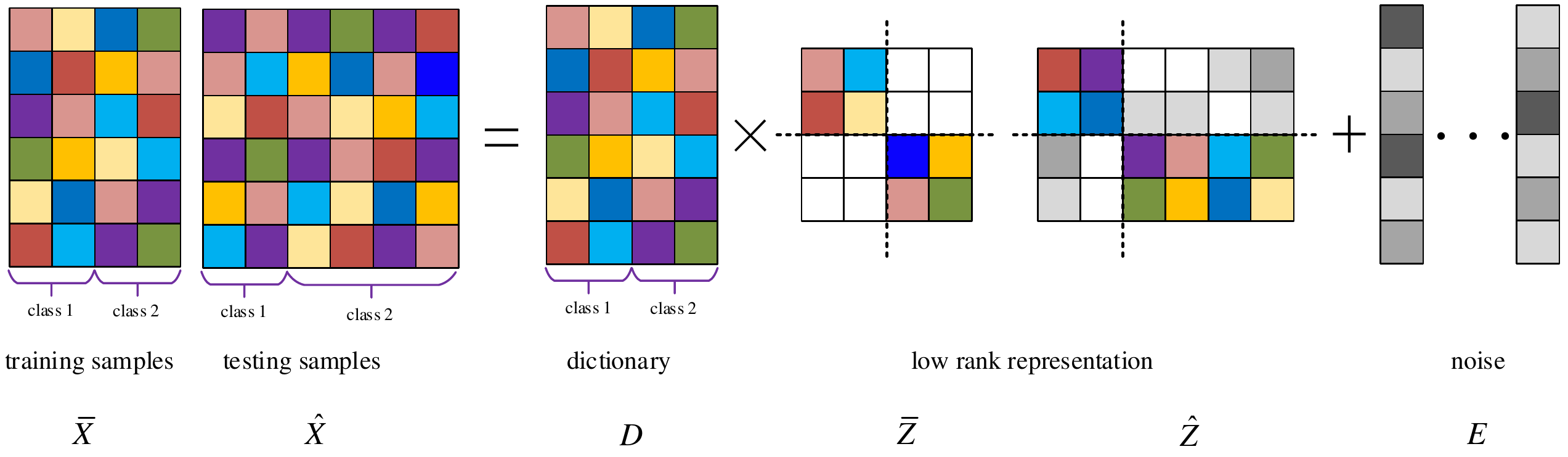,width=17cm}}
	\end{minipage}
	\caption{Illustration of the proposed LSLRR method. The value of colored and white blocks are non-zero and zero, respectively. Besides, the value of gray blocks is close to zero. For the purpose of simplification, only two class are used to describe the method.
	}
	\label{fig:LSLRR}
\end{figure*}

\subsection{Structure Preserving Strategy (SPS) for LSLRR}
\label{ssec:structure constraint}
Hyperspectral data $X$ is first divided into two parts, denoting $X=[\bar{X},\hat{X}]$, where $\bar{X}$ represents the training data and $\hat{X}$ represents the testing data. Rearrange the permutation of samples according to each class that $\bar{X}=[\bar{X}_1,\bar{X}_2,...,\bar{X}_c]\in \mathbb{R}^{d \times m}$, where $X_i$ is the \emph{i}-th class set of training samples, and $c$ denotes the number of classes. Besides, $\hat{X}=[\hat{x}_1,\hat{x}_2,...,\hat{x}_n]\in \mathbb{R}^{d \times n}$ is the testing feature matrix, whose \emph{i}-th column is the spectral vector of the \emph{i}-th testing sample. In LRR model, we set the data $Y=[\bar{X},\hat{X}]$ while the dictionary $A=\bar{X}$. So $[\bar{X},\hat{X}]=\bar{X}Z$ is obtained. Similarly, $Z$ can be written as $[\bar{Z},\hat{Z}]$, where $\bar{Z}$ and $\hat{Z}$ are the low-rank representation for $\bar{X}$ and $\hat{X}$ under the base $\bar{X}$, respectively.

In general LRR model, all data are used as the dictionary and each sample is considered as the atom of the dictionary, e.g. $X=XZ+E$. When removing sparse noise $E$, the data $X$ can be reconstructed by low-rank representation $Z$ based on the data itself. Furthermore, if data samples are permuted based on the order of classes, the ideal representation matrix $Z$ would has a class-wise block-diagonal structure as follows
\begin{equation}
\label{Eq:block-diagonal}
Z=\left[ \begin{matrix}
Z_{1}^{*} & 0 & 0 & 0  \\
0 & Z_{2}^{*} & 0 & 0  \\
0 & 0 & \ddots  & 0  \\
0 & 0 & 0 & Z_{c}^{*}  \\
\end{matrix} \right],
\end{equation}
where $c$ is the number of classes. The proposed model $[\bar{X},\hat{X}]=\bar{X}[\bar{Z},\hat{Z}]+E$ has a similar property to the classical LRR model $X=XZ+E$. That is, representation matrix $\bar{Z}$ and $\hat{Z}$ should also have a class-wise block-dagonal structure as the form of (\ref{Eq:block-diagonal}).

To make $\bar{Z}$ and $\hat{Z}$ hold the above structure, we introduce a structured auxiliary matrix $Q$ to constrain $Z$. Firstly, $Q$ is also divided into two parts: $\bar{Q}$ and $\hat{Q}$. We can obtain $\bar{Z}^*_i$, $i=1,2,...,c$, with setting $A=X_i$ by solving the model (\ref{Eq:LLRR}). Let $\bar{Q}=diag(\bar{Z}^*_1,\bar{Z}^*_2,...,\bar{Z}^*_n)$, where $diag$ is the diagonal operation. Note that this step actually utilizes the label information for the training data $\bar{X}$. So the class-wise block-diagonal structure for $\bar{Z}$ is easy to preserve. Secondly, it's difficult to hold the structure (\ref{Eq:block-diagonal}) for $\hat{Z}$ without a prior about the number of each class testing samples. As is known to us, there're lots of zero elements in $\hat{Z}$ when it has a block-diagonal structure. In addition, we previously mention that $Z_{ij}$ represents the similarity of the \emph{i}-th and \emph{j}-th samples. We employ the Gaussian similarity function to generate the auxiliary matrix $\hat{Q}$ as follows
\begin{align}
	\label{Eq:bar{Q}}
	\hat{Q}_{ij}=exp(-\frac{\lVert x_i-x_j\rVert^2_2+m\lVert l_i-l_j\rVert^2_2}{\sigma}),
\end{align}
where the parameter $\sigma$ is used to control the width of neighbors. If distance between the \emph{i}-th training pixel and the \emph{j}-th testing pixel is large enough (e.g., larger than $\theta$, where $\theta$ is maximum distance parameter), we will set $\lVert x_i-x_j\rVert^2_2+m\lVert l_i-l_j\rVert^2_2=\infty$. Thus, $\hat{Q}$ would has many zeros elements and $\hat{Z}$ would be a sparse matrix. Finally, $Q$ is obtained by $Q=[\bar{Q},\hat{Q}]$. So the structure constraint can be written as $\lVert Z-Q\rVert^2_F$, which makes the low-rank representation $\bar{Z}$ and $\hat{Z}$ have an approximatively block-diagonal structrue.

Considering that the \emph{j}-th column of $Z$ represents the similarity between each training pixels and the \emph{j}-th testing pixel, we enforce the sum of each column of $Z$ to be 1, i.e., $1^T_mZ=1^T_{m+n}$. After incorporating the above two crucial techniques into the classical LRR model, the locality and structure regularized low rank representation can be formulated as
\begin{equation}
\begin{split}
\label{Eq:LSLRR1}
&\mathop{\min}\limits_{Z,E} \quad \lVert Z \rVert_*+\lambda \lVert E \rVert_{2,1}+\alpha \lVert M\circ Z\rVert_1+\beta \lVert Z-Q\rVert^2_F \\
&\text{  } s.t. \quad  X=\bar{X}Z+E,  1^T_mZ=1^T_{m+n}, Z\geq 0,
\end{split}
\end{equation}
where $1_m$ and $1_{m+n}$ are unit vectors with length of $m$ and $m+n$, respectively.

\subsection{Dictionary Learning for LSLRR}
\label{sec:DL}
Dictionary learning is a crucial step for most classification problems. Generally, the whole samples are usually used for the dictionary for LRR. However, when the data samples are corrupted by noise, they can not well reconstruct themselves by polluted dictionary. Besides, high-quality dictionary can improve significantly the performance of classification methods. The process of learning the low rank representation can also become easy with a compact dictionary. Here, we will learn a discriminative dictionary from the corrupted HSI data.

For the problem (\ref{Eq:LSLRR1}), the dictionary is randomly selected from HSI data, and the atoms in $\bar{X}$ are a part of the whole HSI pixels. In the solving process, the dictionary $\bar{X}$ is fixed. However, if the selected samples are not representative and discriminative, or even worse (i.e. grossly corrupted) for the whole data, the obtained low-rank representation $Z$ would be useless. Therefore, we integrate a dictionary learning process into the problem (\ref{Eq:LSLRR1}) instead of fixing some dictionary atoms. Then the final objective function can be demonstrated as
\begin{equation}
\begin{split}
\label{Eq:LSLRR2}
&\mathop{\min}\limits_{Z,E,D} \quad \lVert Z \rVert_*+\lambda \lVert E \rVert_{2,1}+\alpha \lVert M\circ Z\rVert_1+\beta \lVert Z-Q\rVert^2_F \\
&\text{  } s.t. \quad  X=DZ+E, 1^T_mZ=1^T_{m+n}, Z\geq 0,
\end{split}
\end{equation}
where $\alpha$ and $\beta$ control the weights of locality and structure constraints, respectively. The proposed method, namely LSLRR, has a considerable ability to require the block-diagonal representation and simultaneously to learn a discriminative dictionary. In addition, Fig. \ref{fig:LSLRR} illustrates the proposed LSLRR. The given data is first divided into training set $\bar{X}$ and testing set $\hat{X}$. Then the low rank representation matrix $\bar{Z}$ for training set and $\hat{Z}$ for testing set are obtained based on the dictionary $D$. Besides, $\bar{Z}$ is a block-diagonal matrix, and $\hat{Z}$ is an approximately block-diagonal matrix.

\subsection{HSI Classification via LSLRR}
\label{sec:HSI LSLRR}
Hyperspectral pixels belonging to the same class have a extremely similar spectral reflectance curve, which is the theoretical evidence to classify HSI. Although HSI data has a great number of bands and the dimensionality is very high, the similarity between neighboring bands is also very high. \cite{landgrebe2002hyperspectral} indicates that many low-dimensional subspaces exist in HSI data space. Besides, Chakrabarti \emph{et al}. \cite{chakrabarti2011statistics} made a lot of statistical analyses based on real-world HSI data, and came to a conclusion that the rank of HSI data matrix is approximately equal to the number of classes. This implies HSI data satisfy the low-rank property. Pixels of each class have a similar position in the whole HSI space, and they make up a low-dimensional subspace. For the proposed LSLRR, it can effectively segment these subspaces embedded in HSI from both global and local aspects. Recall that $\hat{z}_{ij}$ in $\hat{Z}$ strands for the similarity of the \emph{i}-th training pixel and \emph{j}-th testing pixel. The larger the value of $\hat{z}_{ij}$ is, the higher the possibility of $x_i$ and $x_j$ belongs to the same class. Therefore, the final classification results can be directly obtained and it is no need to employ some complex classification algorithms. Specifically, the label of a testing pixel $x_j$ can be confirmed as follows. First, compute the sum of the \emph{j}-th column of $\hat{Z}$ for each class. The result is denoted by $S_l(\hat{z}_j)$, $l\in [1,...,c]$. Second, the label of $x_j$, denoted by $label(x_j)$, is determined as
\begin{align}
\label{Eq:get label}
label(x_j)=\mathop{\arg\max}\limits_{l=1,...,c} S_l(\hat{z}_j).
\end{align}

\section{Optimization Algorithm for Solving LSLRR}
\label{sec:solution}
In this section, we derive an optimization algorithm to solve the LSLRR model (\ref{Eq:LSLRR2}). In recent years, a great number of algorithms \cite{lin2011linearized}, \cite{lin2010augmented} have been developed to solve the rank minimization optimization problem. Here, we adopt the high-efficiency inexact Augmented Lagrange Multiplier (IALM) method to solve the proposed LSLRR. Firstly, we introduce two auxiliary variables $H$ and $J$ to make the problem (\ref{Eq:LSLRR2}) become easily solvable. Thus, the equivalent problem of (\ref{Eq:LSLRR2}) is converted to
\begin{equation}
\begin{split}
\label{Eq:solve1}
&\mathop{\min}\limits_{H,J,Z,E,D}  \lVert Z \rVert_*+\lambda \lVert E \rVert_{2,1}+\alpha\lVert M\circ Z\rVert_1+\beta \lVert Z-Q\rVert^2_F \\
&\quad  s.t. \quad   X=DZ+E,Z=J,H=Z, 1^T_mZ=1^T_{m+n}, Z\geq 0.
\end{split}
\end{equation}
Then the corresponding augmented Lagrangian function for (\ref{Eq:solve1}) can be written as
\begin{equation}
\begin{split}
\label{Eq:solve2}
&\mathop{\min}\limits_{H\geq 0,J,Z,E,D}  \lVert J \rVert_*+\lambda \lVert E \rVert_{2,1}+\alpha\lVert M\circ H\rVert_1+\beta \lVert Z-Q\rVert^2_F \\
&+<Y_1,X-DZ-E>+<Y_2,Z-J>+<Y_3,H-Z>\\
&+<Y_4,1^T_mZ-1^T_{m+n}>+\frac{\mu}{2}(\lVert X-DZ-E\rVert^2_F+\lVert Z-J\rVert^2_F\\
&+\lVert H-Z\rVert^2_F+\lVert 1^T_mZ-1^T_{m+n}\rVert^2_F),
\end{split}
\end{equation}
where $<A,B>=trace(A^TB)$, $\mu >0$ is a penalty parameter and $Y_1$, $Y_2$, $Y_3$ and $Y_4$ are Lagrange multipliers. The alternative optimization algorithm can be applied to solve the problem (\ref{Eq:solve2}) with five optimization variables ($H, J, Z, E, D$). The detailed updating schemes can be seen as follows.

\textbf{Updata H}: fix $J$, $Z$, $E$, and $D$, and then $H$ can be updated as follows
\begin{equation}
\begin{split}
\label{Eq:update H}
H^{k+1}=\mathop{\arg\min}_{H\geq 0} \frac{\alpha}{\mu ^k} \lVert M\circ H^k\rVert_1+\frac{1}{2}\lVert H^k-Z^k+\frac{Y^k_3}{\mu ^k} \rVert ^2_F.
\end{split}
\end{equation}
The solution for (\ref{Eq:update H}) can be computed \cite{tang2014structure} by
\begin{align}
\label{Eq:solve H}
H^{k+1}_{ij}=max[0,\Theta _{w_{ij}}(Z^k_{ij}-\frac{Y^k_{3,ij}}{\mu^k})],
\end{align}
where $\Theta _w(x)=max(x-w,0)+min(x+w,0)$, $w_{ij}=(\alpha /\mu^k)M_{ij}$.

\textbf{Updata J}: fix $H$, $Z$, $E$, and $D$, and then $J$ can be updated as follows
\begin{equation}
\begin{split}
\label{Eq:update J}
&J^{k+1}=\mathop{\arg\min}_J \frac{1}{\mu^k}\lVert J^k\rVert_*+\frac{1}{2}\lVert Z^k-J^k+\frac{Y^k_2}{\mu^k}\rVert ^2_F\\
&\quad\quad=US_{1/\mu^k}(\Sigma)V^T,
\end{split}
\end{equation}
where $U\Sigma V^T$ is the singular value decomposition (SVD) of $Z^k+Y^k_2/\mu^k$, and $S_\epsilon(x)=sgn(x)max(|x|-\epsilon,0)$ is the soft-thresholding operator \cite{liu2010robust}.

\textbf{Updata Z}: fix $H$, $J$, $E$, and $D$, and then $Z$ can be updated as follows
\begin{equation}
\begin{split}
\label{Eq:update Z}
&Z^{k+1}=\mathop{\arg\min}_Z \beta\lVert Z^k-Q\rVert^2_F+\frac{\mu^k}{2}\lVert Z^k-J^k+\frac{Y^k_2}{\mu ^k}\rVert ^2_F \\
&+\frac{\mu^k}{2}\lVert X-DZ^k-E^k+\frac{Y^k_1}{\mu^k}\rVert ^2_F+\frac{\mu^k}{2}\lVert H^k-Z^k+\frac{Y^k_3}{\mu^k}\rVert ^2_F\\
&+\frac{\mu^k}{2}\lVert 1^T_mZ^k-1^T_{m+n}+\frac{Y^k_4}{\mu^k}\rVert^2_F.
\end{split}
\end{equation}
Problem (\ref{Eq:update Z}) is a quadratic minimization problem. And it has a closed-form solution, which can be obtained by making the derivative of (\ref{Eq:update Z}) be zero. The optimal solution for variable $Z$ is
\begin{equation}
\begin{split}
\label{Eq:solve Z}
Z^{k+1}=[W^k]^{-1}[2\beta Q^k+\mu^k(D^TA^k+B^k+C^k+1_mF^k)],
\end{split}
\end{equation}
where $A=X-E+Y_1/\mu$, $B=J-Y_2/\mu$, $C=H+Y_3/\mu$, $F=1^T_{m+n}-Y_4/\mu$, and $W=2\beta I+\mu(D^TD+2I+1_m1^T_m)$.

\textbf{Updata E}: fix $H$, $J$, $Z$, and $D$, and then $E$ can be updated as follows
\begin{align}
\label{Eq:update E}
E^{k+1}=\mathop{\arg\min}_E \frac{\lambda}{\mu^k}\lVert E^k\rVert_{2,1}+\frac{1}{2}\lVert X-DZ^k-E^k+\frac{Y^k_1}{\mu^k}\rVert ^2_F.
\end{align}
Denote $G=X-DZ+Y_1/\mu$, then the \emph{j}-th column of optimal $E$ \cite{liu2010robust} is
\begin{equation}
\label{Eq:solve E}
{{E}^{k+1}}(:,i)=\left\{ \begin{matrix}
\frac{||{{g}_{i}}|{{|}_{2}}-\frac{\lambda }{{{\mu }^{k}}}}{||{{g}_{i}}|{{|}_{2}}}{{g}_{i}},\text{  }if\text{ }\frac{\lambda }{{{\mu }^{k}}}<||{{g}_{i}}|{{|}_{2}},  \\
0,\text{               }otherwise.  \\
\end{matrix} \right.
\end{equation}

\textbf{Updata D}: fix $H$, $J$, $Z$, and $E$, and then $D$ can be updated as follows
\begin{align}
\label{Eq:update D}
D^{k+1}=\mathop{\arg\min}_D \frac{\mu^k}{2}\lVert X-D^kZ^k-E^k+\frac{Y^k_1}{\mu^k}\rVert ^2_F.
\end{align}
Problem (\ref{Eq:update D}) is also a quadratic minimization problem. Here, we employ an iteration updating strategy to obtain the optimal solution of dictionary $D$. Firstly, we initialize the dictionary $D^0$ by randomly selecting a part of HSI pixels. Secondly, the updating dictionary $D^{new}$ is obtained by solving the problem (\ref{Eq:update D}). Finally, the detailed updating rule is
\begin{align}
	\label{Eq:solve D}
	D^{k+1}=wD^k+(1-w)D^{new},
\end{align}
where $w$ is a weight parameter. For each iteration, $D^{new}=(X-E+Y^k_1/\mu^k)Z^T(ZZ^T)^{-1}$.

Finally, the overall optimization algorithm for solving the proposed LSLRR (\ref{Eq:LSLRR2}) is described as Algorithm \ref{Alg:IALM}.

\begin{algorithm}
	\caption{IALM for solving LSLRR}
	
	\hspace*{0.02in} {\bf Input:}
	testing set $\hat{X}$, training set $\bar{X}$, local constraint matrix $M$, structure constraint matrix $Q$, parameter $\lambda$, $\alpha$, $\beta$, $m$.
	
	\hspace*{0.02in} {\bf Output:}
	low rank represeentation $Z$, the noise $E$.
	
	\hspace*{0.02in} {\bf Initialize:}
	$H=J=Z=E=0$, $D^0=\bar{X}$, $\mu=10^{-6}$, $max_{\mu}=10^{10}$, $\rho=1.1$, $\varepsilon=10^{-4}$, $Y_1=Y_2=Y_3=Y_4=0$.
	
	\hspace*{0.02in} {\bf While} not converged \textbf{do}
	\begin{enumerate}
		\item Compute the optimal solution of $H$, $J$, $Z$, $E$ and $D$ according to (\ref{Eq:solve H}), (\ref{Eq:update J}), (\ref{Eq:solve Z}), (\ref{Eq:solve E}), (\ref{Eq:solve D}), respectively.
		\item Update the Lagrange multipliers by
		
		$Y^{k+1}_1=Y^k_1+\mu^k(X-\bar{X}Z^k-E^k)$,
		
		$Y^{k+1}_2=Y^k_2+\mu^k(Z^k-J^k)$,
		
		$Y^{k+1}_3=Y^k_3+\mu^k(H^k-Z^k)$,
		
		$Y^{k+1}_4=Y^k_4+\mu^k(1^T_mZ^k-1^T_{m+n})$.
		\item Update the parameter $\mu$ by $\mu^{k+1}=max(\rho \mu^k,max_{\mu})$.
		\item Check the convergence conditions
		
		$\lVert X-DZ-E\rVert_\infty<\varepsilon$, $\lVert Z-J\rVert_\infty<\varepsilon$,
		
		$\lVert H-Z\rVert_\infty<\varepsilon$, $\lVert D^{k+1}-D^{k}\rVert_\infty<\varepsilon$,
		
		$\lVert 1^T_mZ-1^T_{m+n}\rVert_\infty<\varepsilon$.
		\item $k \gets k+1$.
	\end{enumerate}
	\hspace*{0.02in} {\bf End while}
	
	\label{Alg:IALM}
\end{algorithm}


\begin{figure*}[htb]
\centering
  \begin{tabular}{@{}cccc@{}}
    \includegraphics[width=5.5in]{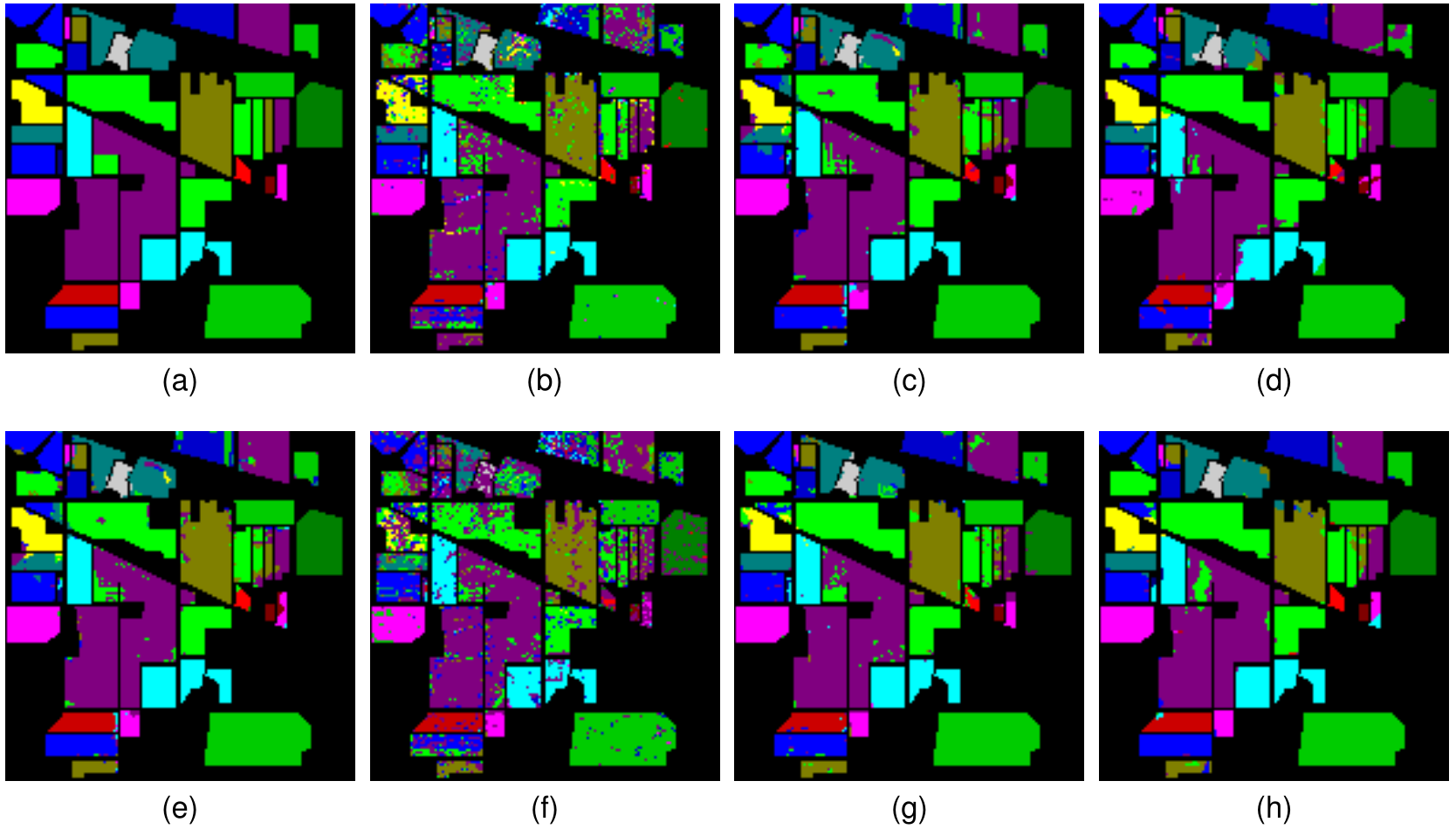} &
    \includegraphics[width=1.2in]{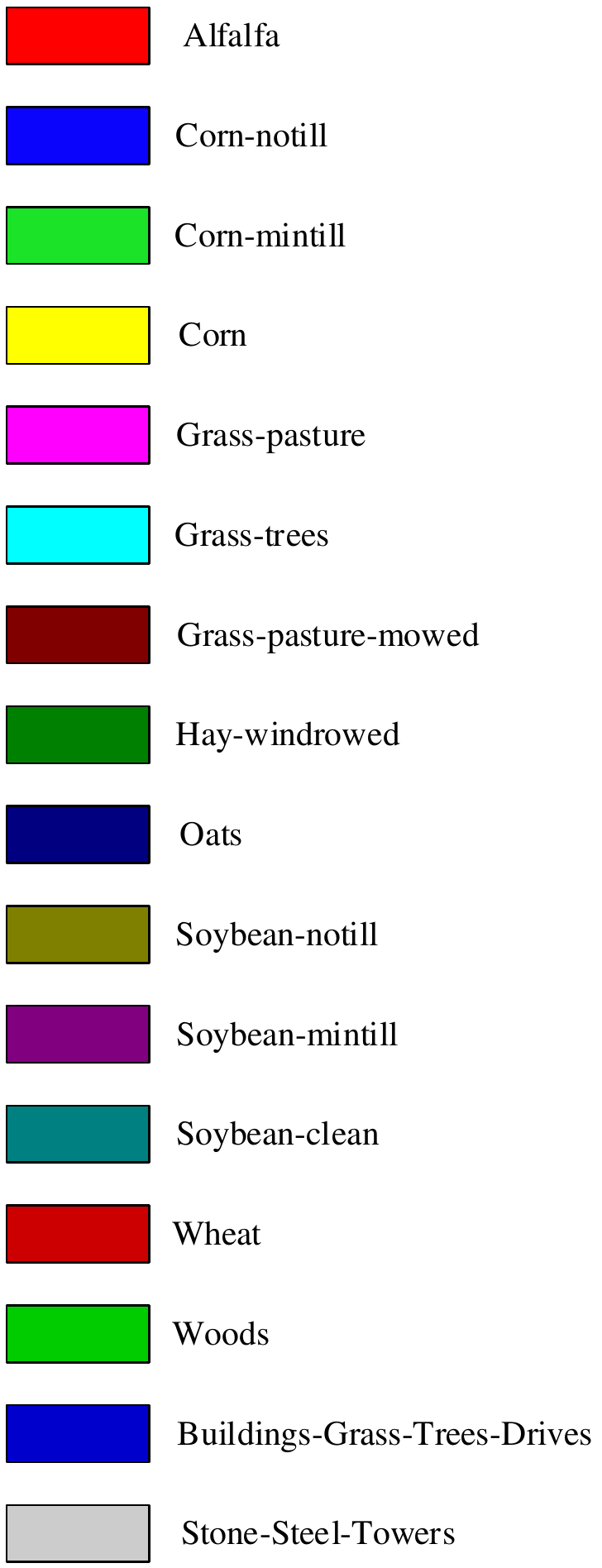}
  \end{tabular}
  \caption{Ground truth and classification maps for Indian Pines. (a) Ground truth, (b) SVM (81.67\%), (c) SVMCK (91.93\%), (d) JRSRC (92.36\%), (e) cdSRC (93.61\%), (f). LRR (70.47\%), (g) LGIDL (94.52\%), (h) LSLRR (95.63\%).}
  \label{fig:inp_maps}
\end{figure*}



\begin{figure*}[htb]
\centering
  \begin{tabular}{ll}
    \adjustbox{valign=t}{\includegraphics[width=5.5in]{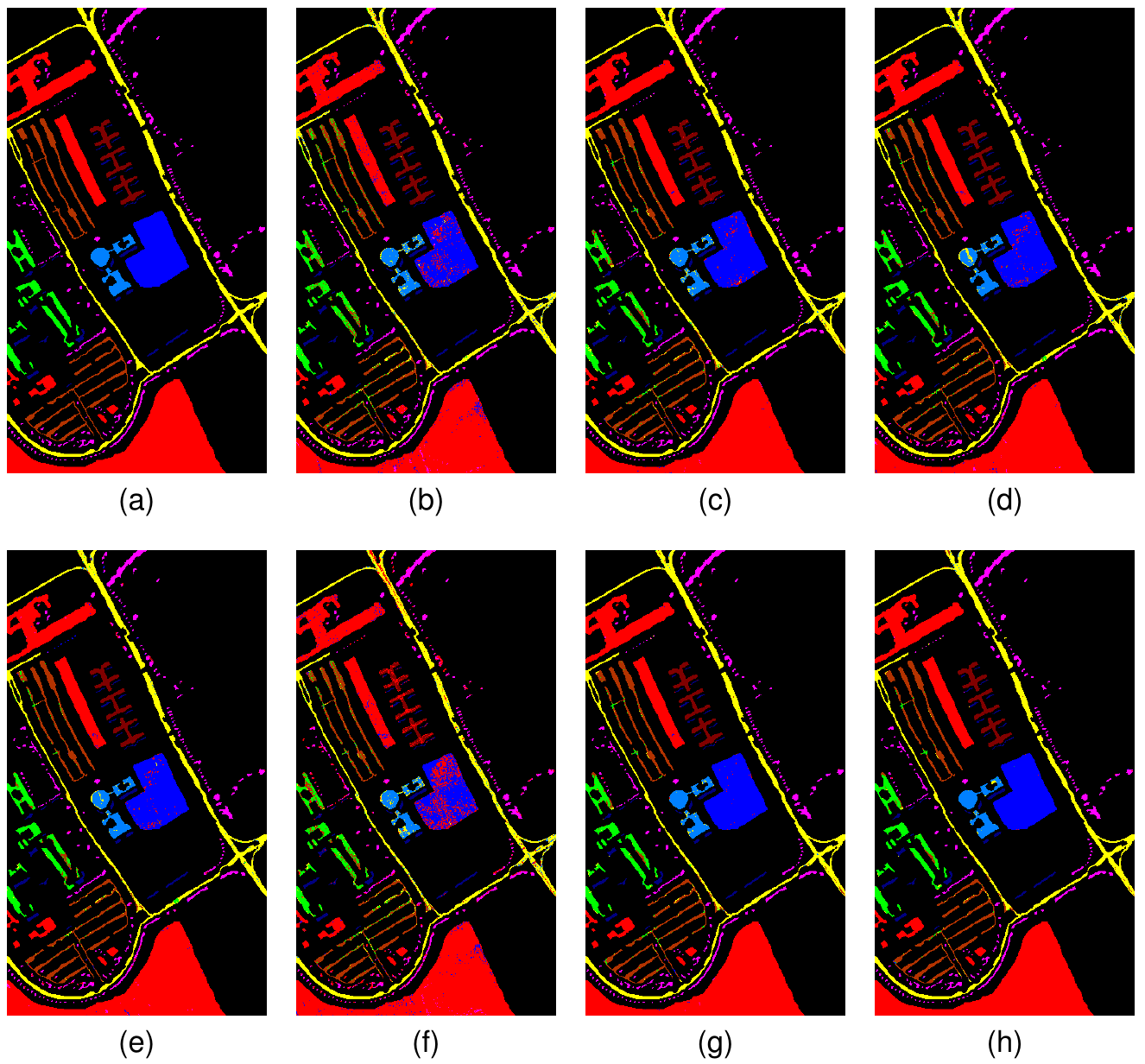}} &
    \adjustbox{valign=t}{\includegraphics[width=1.2in]{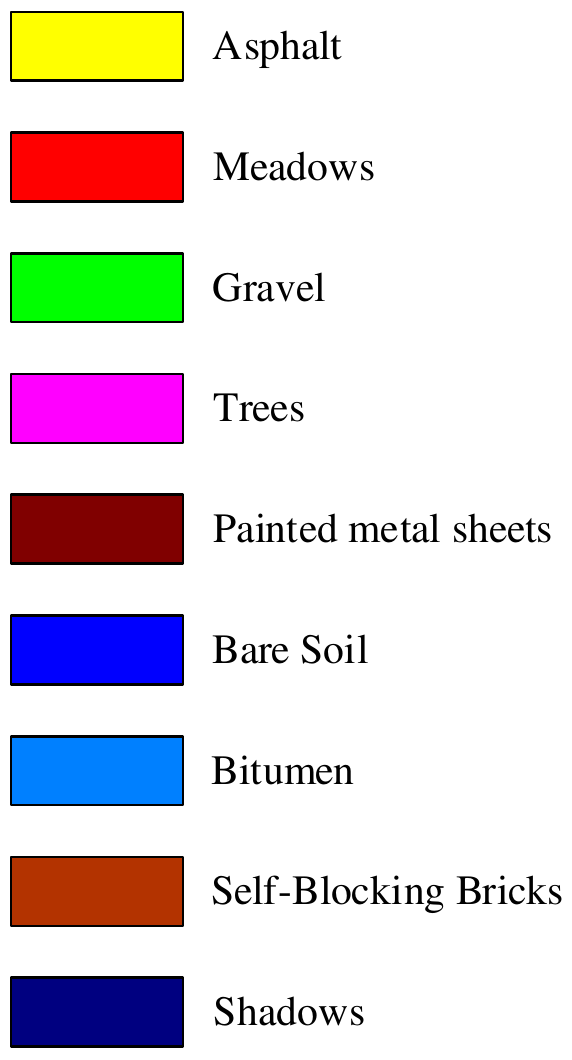}}
  \end{tabular}
  \caption{Ground truth and classification maps for Pavia University. (a) Ground truth, (b) SVM (93.05\%), (c) SVMCK (95.86\%), (d) JRSRC (96.24\%), (e) cdSRC (97.02\%), (f). LRR (86.73\%), (g) LGIDL (97.81\%), (h) LSLRR (98.52\%).}
  \label{fig:pav_maps}
\end{figure*}


\begin{figure*}[htb]
\centering
  \begin{tabular}{ll}
    \adjustbox{valign=t}{\includegraphics[width=5.5in]{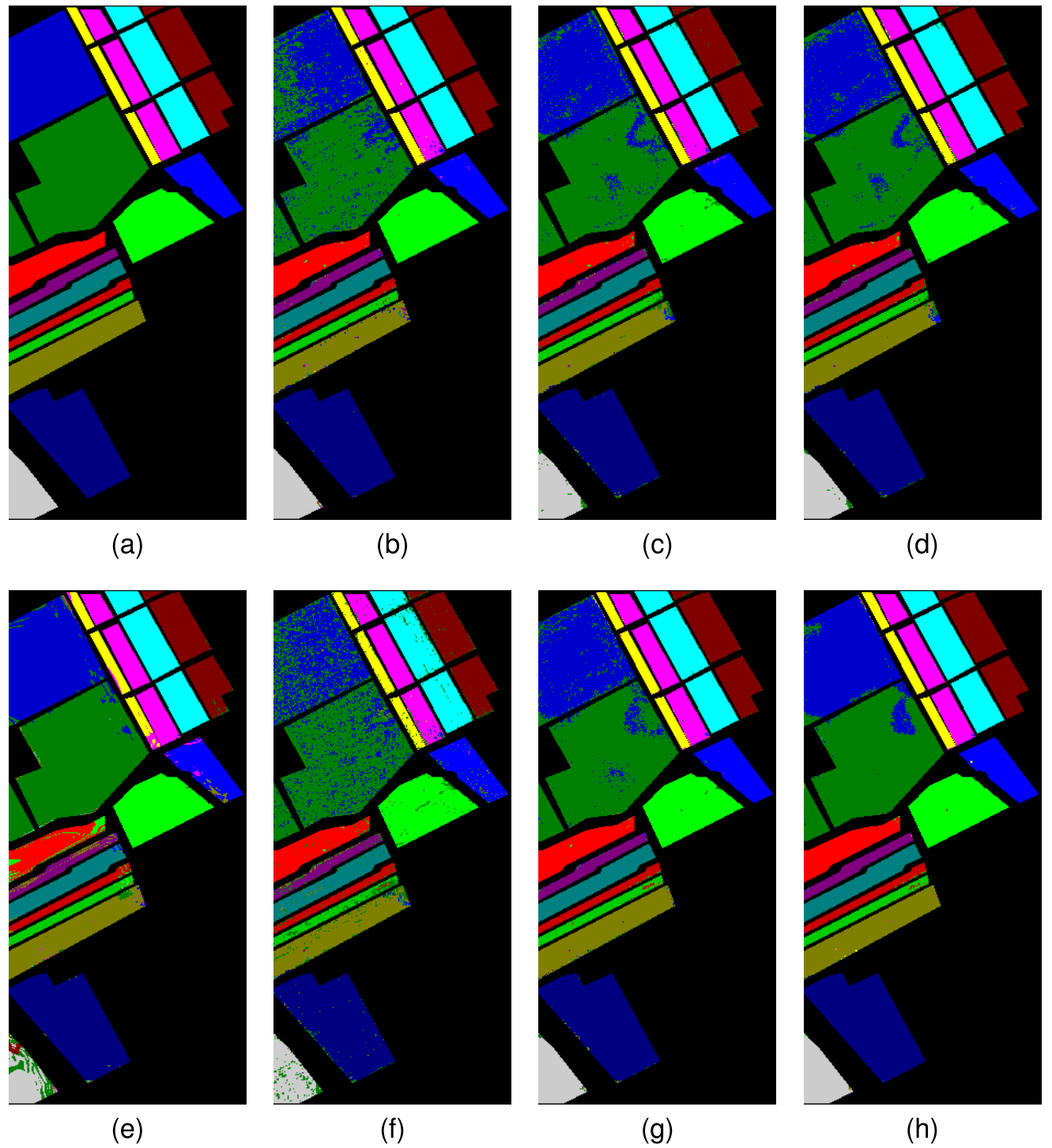}} &
    \adjustbox{valign=t}{\includegraphics[width=1.2in]{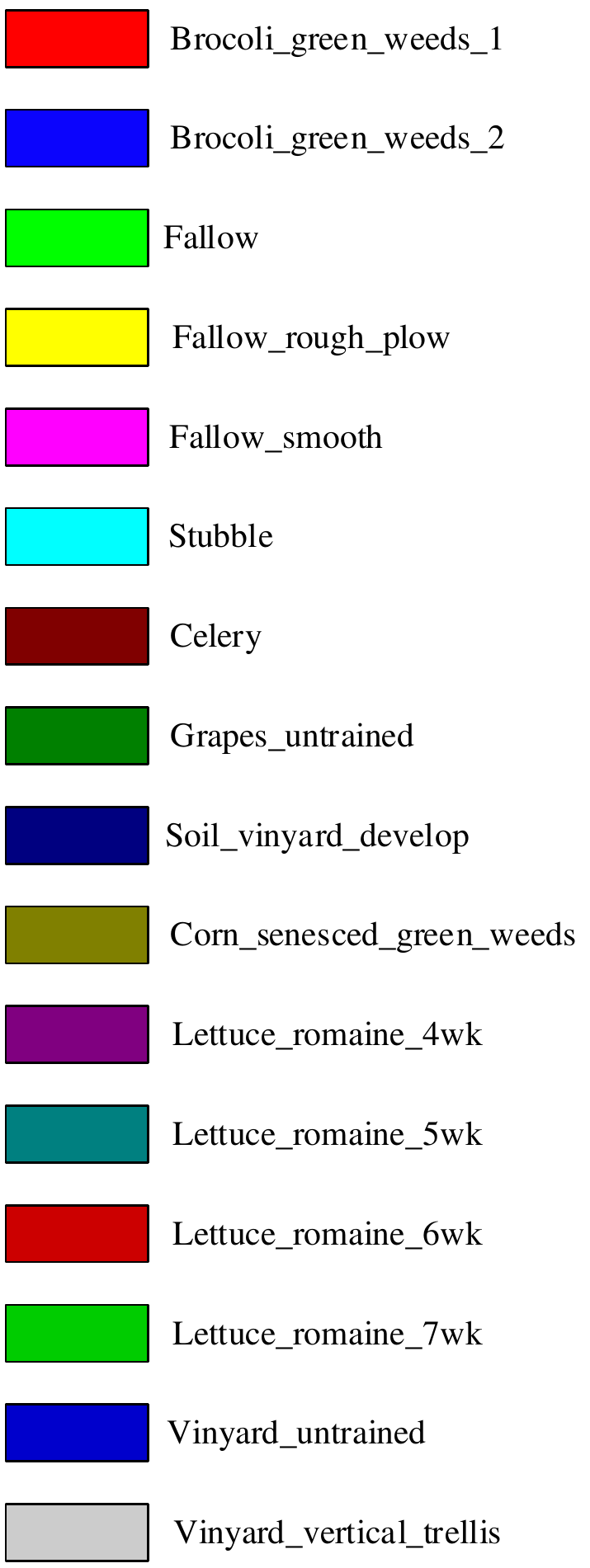}}
  \end{tabular}
  \caption{Ground truth and classification maps for Salinas. (a) Ground truth, (b) SVM (92.64\%), (c) SVMCK (95.46\%), (d) JRSRC (95.84\%), (e) cdSRC (96.13\%), (f). LRR (86.81\%), (g) LGIDL (96.58\%), (h) LSLRR (97.77\%).}
  \label{fig:sal_maps}
\end{figure*}




\renewcommand\arraystretch{1.2}
\begin{table*}[htb]
	\centering
	\caption{Classification accuracy (\%) of different comparison methods and the proposed lslrr for indian pines dataset}
	\begin{tabular}{|p{1.2cm}<{\centering}||p{1.2cm}<{\centering}|p{1.2cm}<{\centering}|p{1.2cm}<{\centering}|p{1.2cm}<{\centering}|p{1.2cm}<{\centering}|p{1.2cm}<{\centering}|p{1.2cm}<{\centering}|}
		\hline
		Class & SVM   & SVMCK & JRSRC & cdSRC & LRR   & LGIDL & LSLRR \\\hline\hline
		1     & 87.80 & 70.73 & 58.54 & 85.37 & 19.15 & 63.41 & \textbf{100}   \\\hline
		2     & 78.29 & 88.17 & 92.68 & 91.05 & 63.04 & \textbf{96.26} & 95.69 \\\hline
		3     & 64.66 & 88.35 & \textbf{95.18} & 91.16 & 56.36 & 91.97 & 94.25 \\\hline
		4     & 77.46 & 91.08 & 92.96 & 94.84 & 21.13 & 87.32 & \textbf{98.05} \\\hline
		5     & 91.72 & 88.74 & 88.51 & 92.18 & 75.17 & 90.80 & \textbf{94.42} \\\hline
		6     & 97.41 & 97.72 & 87.21 & \textbf{99.39} & 86.15 & 99.09 & 98.93 \\\hline
		7     & 64.02 & 100   & 72.00 & 100   & 47.97 & 84.01 & \textbf{100}   \\\hline
		8     & 98.14 & 98.14 & 99.07 & 100   & 79.53 & 96.98 & \textbf{100}   \\\hline
		9     & 33.33 & 38.89 & 33.33 & \textbf{50.00} & 11.11 & 38.89 & 27.78 \\\hline
		10    & 70.15 & 88.23 & 84.91 & 89.83 & 72.11 & 90.17 & \textbf{90.37} \\\hline
		11    & 83.52 & 96.06 & \textbf{97.56} & 95.97 & 83.88 & 95.74 & 95.79 \\\hline
		12    & 66.88 & 81.84 & 82.02 & 85.39 & 37.45 & 90.26 & \textbf{95.51} \\\hline
		13    & 95.65 & 90.76 & 88.04 & 94.57 & 80.43 & 94.02 & \textbf{96.20} \\\hline
		14    & 94.82 & 98.86 & 95.69 & 98.95 & 90.86 & 99.21 & \textbf{99.56} \\\hline
		15    & 59.37 & 81.27 & 95.10 & 82.42 & 24.50 & 94.24 & \textbf{100}   \\\hline
		16    & 94.05 & 89.29 & 80.95 & 94.05 & 17.86 & 89.29 & \textbf{97.42} \\\hline\hline
		\textbf{OA}    & 81.67 & 91.93 & 92.36 & 93.61 & 70.47 & 94.52 & \textbf{95.63} \\\hline
		\textbf{AA}    & 78.60 & 86.76 & 83.99 & 90.32 & 54.19 & 87.60 & \textbf{92.74} \\\hline
		\textbf{kappa} & 0.7902 & 0.9076 & 0.9124 & 0.9270 & 0.6545 & 0.9374 & \textbf{0.9512} \\\hline
	\end{tabular}
	\label{tab:inp}
\end{table*}

\renewcommand\arraystretch{1.2}
\begin{table*}[htb]
	\centering
	\caption{Classification accuracy (\%) of different comparison methods and the proposed lslrr for pavia university dataset}
	\begin{tabular}{|p{1.2cm}<{\centering}||p{1.2cm}<{\centering}|p{1.2cm}<{\centering}|p{1.2cm}<{\centering}|p{1.2cm}<{\centering}|p{1.2cm}<{\centering}|p{1.2cm}<{\centering}|p{1.2cm}<{\centering}|}
		\hline
		Class & SVM   & SVMCK & JRSRC & cdSRC & LRR   & LGIDL & LSLRR \\\hline\hline
		1     & 94.11 & 96.02 & 95.62 & 96.57 & 88.46 & 96.81 & \textbf{97.33} \\\hline
		2     & 96.94 & 99.63 & 99.26 & 99.44 & 97.06 & 99.79 & \textbf{99.98} \\\hline
		3     & 81.44 & 82.40 & 88.82 & 89.27 & 72.67 & 89.22 & \textbf{91.98} \\\hline
		4     & 94.37 & 97.32 & 91.69 & 93.27 & 74.41 & 98.18 & \textbf{98.73} \\\hline
		5     & 99.30 & 97.03 & 99.84 & 99.92 & 68.47 & 100   & \textbf{100}   \\\hline
		6     & 86.73 & 95.63 & 94.91 & 96.19 & 67.54 & 99.35 & \textbf{99.90} \\\hline
		7     & 86.30 & 89.47 & 87.89 & 92.64 & 80.36 & 94.70 & \textbf{96.52} \\\hline
		8     & 84.02 & 91.14 & 92.68 & 94.08 & 82.68 & 92.17 & \textbf{94.97} \\\hline
		9     & 99.89 & 98.00 & 99.59 & \textbf{99.89} & 94.56 & 98.89 & 99.11 \\\hline\hline
		\textbf{OA}    & 93.05 & 95.86 & 96.24 & 97.02 & 86.73 & 97.81 & \textbf{98.52} \\\hline
		\textbf{AA}    & 91.46 & 94.07 & 94.47 & 95.70 & 80.69 & 96.57 & \textbf{97.61} \\\hline
		\textbf{kappa} & 0.9078 & 0.9524 & 0.9499 & 0.9605 & 0.8186 & 0.9710 & \textbf{0.9804} \\\hline
	\end{tabular}
	\label{tab:pav}
\end{table*}

\renewcommand\arraystretch{1.2}
\begin{table*}[htb]
	\centering
	\caption{Classification accuracy (\%) of different comparison methods and the proposed lslrr for salinas dataset}
	\begin{tabular}{|p{1.2cm}<{\centering}||p{1.2cm}<{\centering}|p{1.2cm}<{\centering}|p{1.2cm}<{\centering}|p{1.2cm}<{\centering}|p{1.2cm}<{\centering}|p{1.2cm}<{\centering}|p{1.2cm}<{\centering}|}
		\hline
		Class & SVM   & SVMCK & JRSRC & cdSRC & LRR   & LGIDL & LSLRR \\\hline\hline
		1     & 99.32 & 98.59 & 98.48 & 88.24 & 96.12 & 99.53 & \textbf{99.69} \\\hline
		2     & 99.87 & 98.14 & 98.81 & 99.86 & 96.07 & 99.46 & \textbf{99.95} \\\hline
		3     & 99.52 & 99.04 & 99.25 & 91.21 & 94.25 & 99.73 & \textbf{99.73} \\\hline
		4     & 98.79 & \textbf{99.40} & 98.04 & 87.31 & 92.37 & 99.09 & 99.17 \\\hline
		5     & 97.76 & 97.41 & 97.44 & \textbf{99.65} & 93.87 & 98.86 & 99.06 \\\hline
		6     & 99.67 & 98.94 & 98.75 & 99.79 & 94.42 & 99.60 & \textbf{99.79} \\\hline
		7     & 99.57 & 98.44 & 99.26 & 98.71 & 96.62 & 99.50 & \textbf{99.67} \\\hline
		8     & 88.35 & 92.82 & 93.59 & \textbf{97.64} & 82.73 & 93.99 & 94.27 \\\hline
		9     & 99.86 & 99.02 & 99.44 & 99.63 & 97.88 & 99.41 & \textbf{99.92} \\\hline
		10    & 95.41 & 93.38 & 94.84 & 96.08 & 87.32 & 97.85 & \textbf{99.10} \\\hline
		11    & 96.45 & 92.32 & 95.86 & 78.23 & 79.70 & 96.75 & \textbf{99.21} \\\hline
		12    & 99.67 & 99.73 & 100   & 97.87 & 94.81 & 99.95 & \textbf{100}   \\\hline
		13    & 97.74 & 96.21 & 94.48 & 93.68 & 92.64 & 97.82 & \textbf{98.01} \\\hline
		14    & \textbf{96.95} & 92.03 & 94.39 & 91.73 & 72.44 & 92.62 & 94.09 \\\hline
		15    & 68.79 & 88.94 & 88.20 & \textbf{96.09} & 61.87 & 89.04 & 95.10 \\\hline
		16    & 99.30 & 96.45 & 96.97 & 77.73 & 86.49 & 98.66 & \textbf{99.42} \\\hline\hline
		\textbf{OA}    & 92.64 & 95.46 & 95.84 & 96.13 & 86.81 & 96.58 & \textbf{97.77} \\\hline
		\textbf{AA}    & 96.06 & 96.30 & 96.74 & 93.34 & 88.72 & 97.62 & \textbf{98.51} \\\hline
		\textbf{kappa} & 0.9179 & 0.9494 & 0.9536 & 0.9524 & 0.8520 & 0.9619 & \textbf{0.9752} \\\hline
	\end{tabular}
	\label{tab:sal}
\end{table*}

\begin{figure*}[htb]
	\centering
	\includegraphics[width=7in]{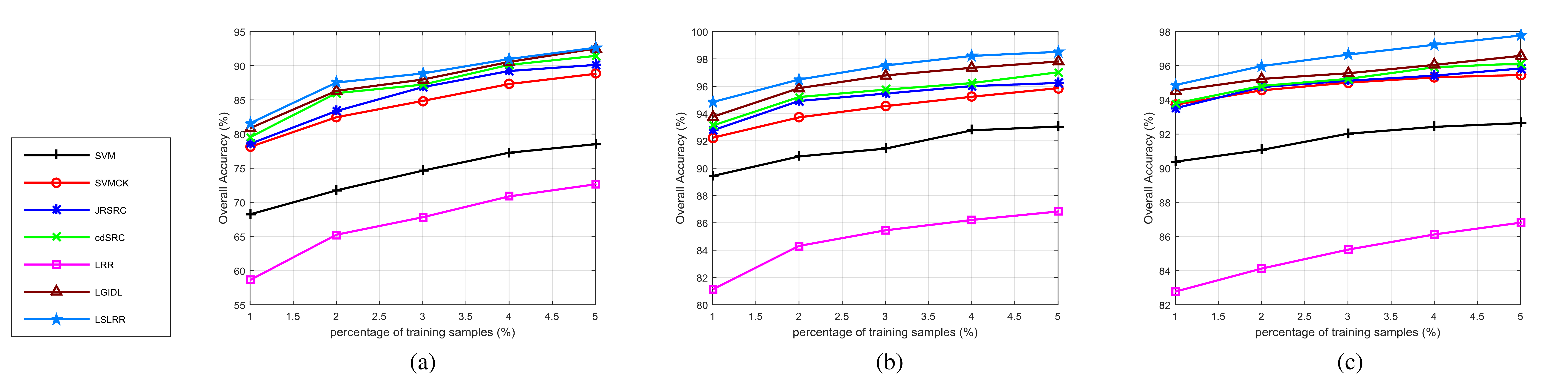}
	\caption{Comparison of overall accuracy for all methods under different percentage of training samples. (a) Indian Pines, (b) Pavia University, (c) Salinas.}
	\label{fig:percent}	
\end{figure*}


\begin{figure}[htb]
	\centering
	\includegraphics[width=3.1in]{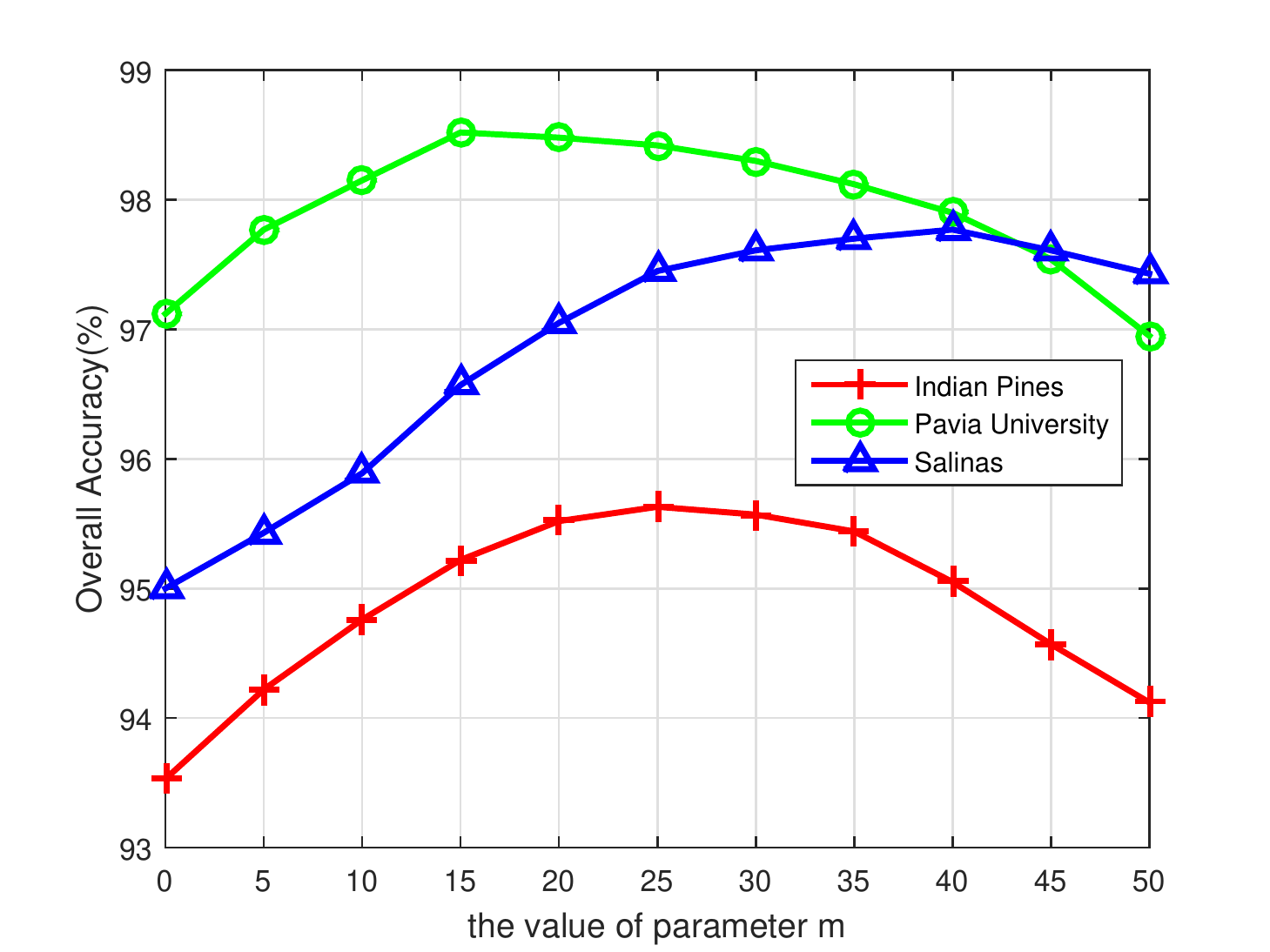}
	\caption{Overall accuracy of three HSI datases under different values of parameter $m$.}
	\label{fig:para_m}
\end{figure}
%
%
%
%
\begin{figure}[htb]
	\centering
	\includegraphics[width=3.1in]{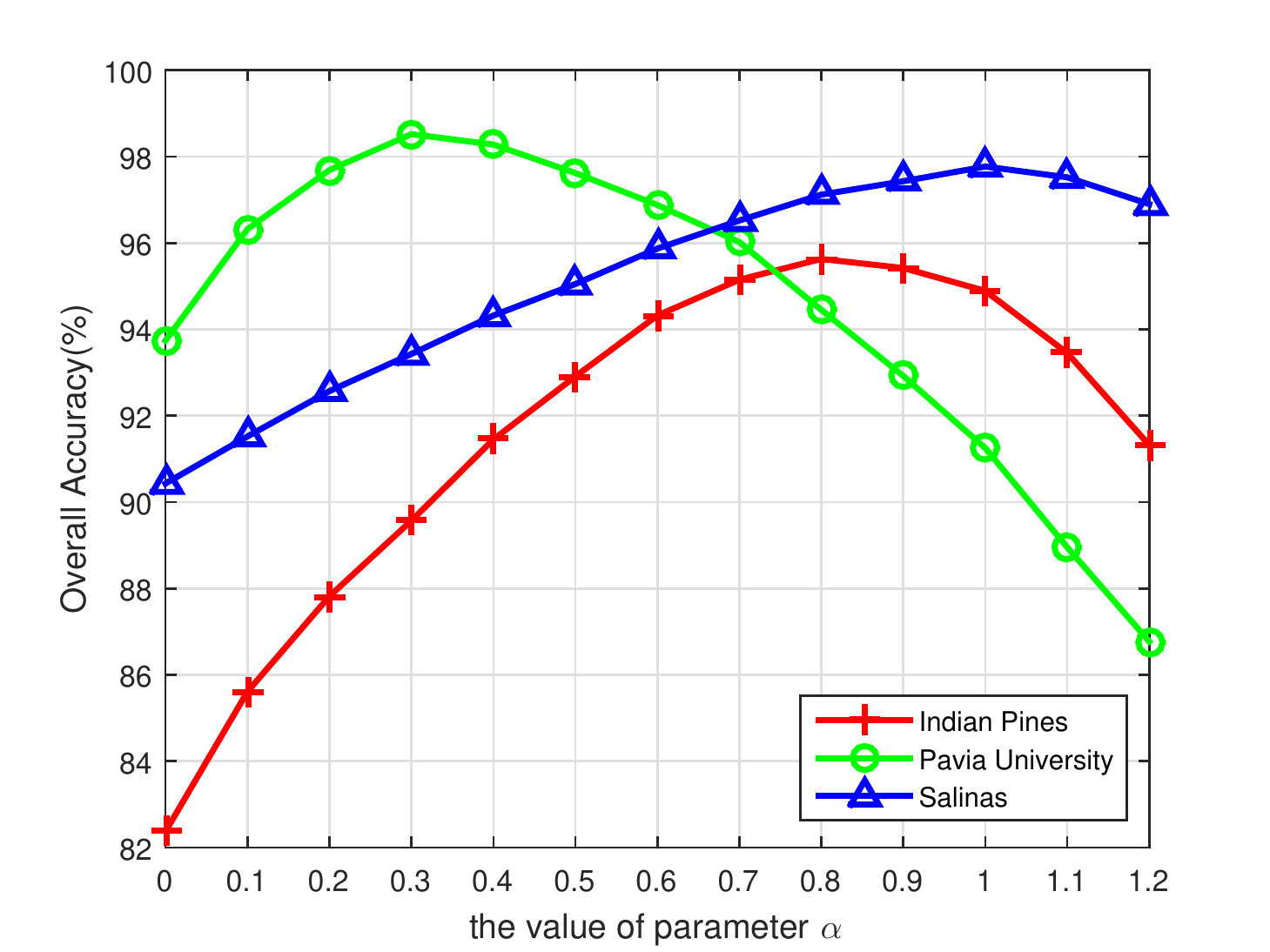}
	\caption{Overall accuracy of three HSI datases under different values of parameter $\alpha$.}
	\label{fig:para_alpha}
\end{figure}
\begin{figure}[htb]
	\centering
	\includegraphics[width=3.1in]{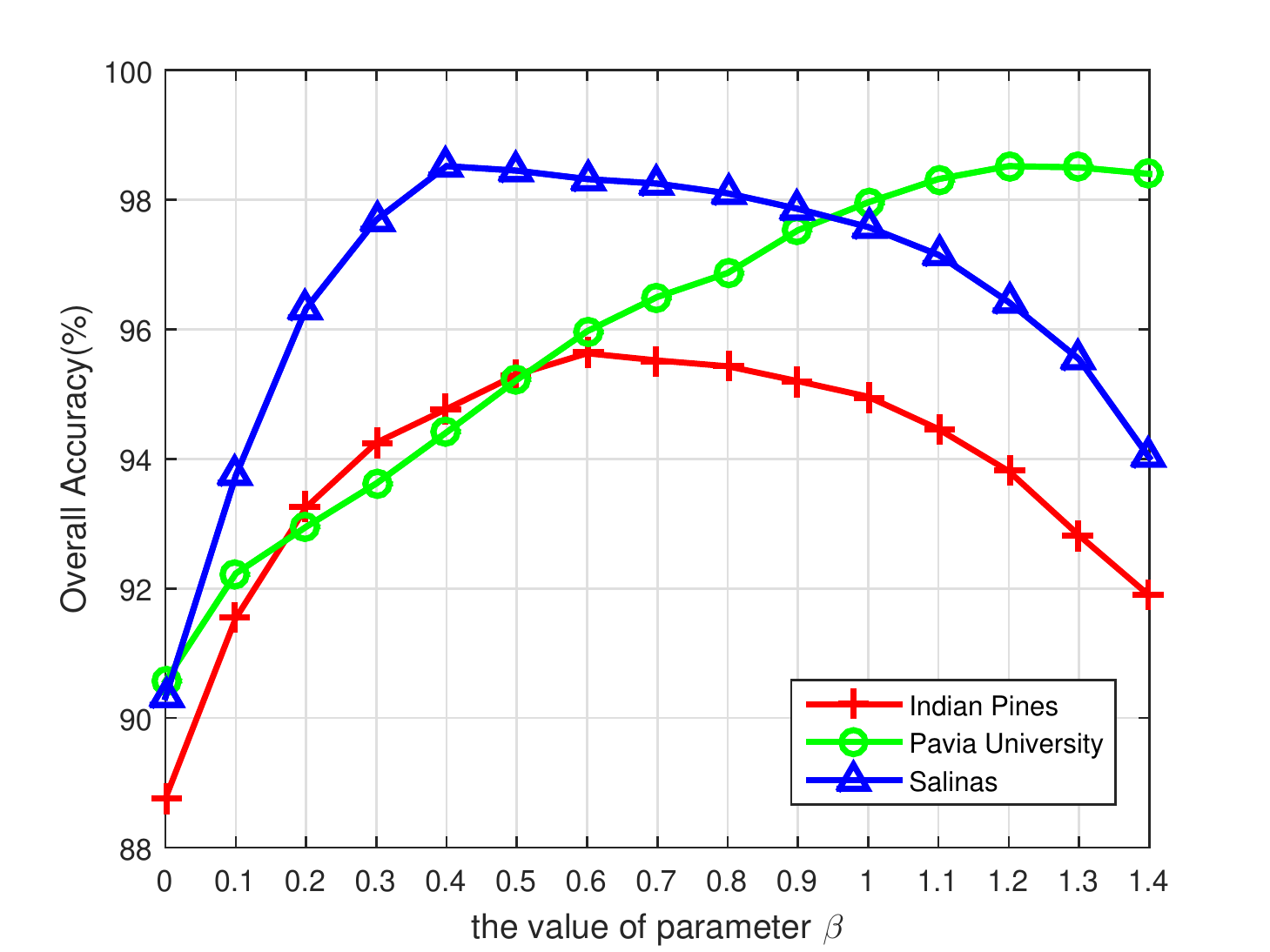}
	\caption{Overall accuracy of three HSI datases under different values of parameter $\beta$.}
	\label{fig:para_beta}	
\end{figure}

\section{Experiments and Analyses}
\label{sec:experiments}
In this section, some comprehensive experiments are conducted to prove the effectiveness of the proposed LSLRR for HSI classification. Many state-of-the-art classification algorithms are considered as the comparison methods. After the experiments, some detailed analyses are also given.

\subsection{Dataset Descriptions}
To evaluate the classification performance of the proposed LSLRR model, three popular hyperspectral datasets are used to conduct the verification experiments. The detailed descriptions are shown as follows \cite{hyper_dataset}.
\begin{enumerate}
	\item \textbf{Indian Pines}: The scene is collected by AVIRIS sensor over the most agricultural regions in the northwestern Indiana, America. And the dataset is composed of $145\times 145$ pixels with 220 spectral bands whose wavelength ranges from 0.4-2.5 $\mu m$. After removing some noise and water-absorption bands, the remaining image has 200 spectral bands, which can be used for classification task. In addition, there are 16 classes for this dataset.
	\item \textbf{Pavia University}: This dataset was captured by ROSIS sensor over the urban area of the University of Pavia, northern Italy, on July 8, 2002. The original dataset consists of 115 spectral bands covering 0.43-0.86 $\mu m$, of which 12 noisy bands are removed and 103 bands are retained. The size of each band is $610\times 340$ with a spatial resolution of 1.3 meters per pixel. Nine categories of ground covering are considered for the classification experiments.
	\item \textbf{Salinas}: The image is also gathered by AVIRIS sensor and contains the wavelength range of 0.4-2.5 $\mu m$ like the Indian Pines. It has a high spatial resolution of 3.7 meters per pixel. The covered area consists of 512 lines and 217 samples. Besides, there are 204 spectral bands after discarding some polluted bands. The number of ground category is also 16. This scene mainly consists of bare soils, vegetables, and vineyard fields.
\end{enumerate}

\subsection{Experimental Setups}
\label{ssec:Experimental Setups}
Before demonstrating the experimental results, the comparison methods, corresponding parameter settings and evaluation indexes are first introduced as follows.

\textbf{1). Comparison Algorithms}: To verify the superiority of the proposed LSLRR, some state-of-the-art HSI classification methods are considered. They are 1) SVM \cite{SVM_compare}; 2) SVMCK \cite{SVMCK_compare}; 3) JRSRC \cite{JRSRC_compare}; 4) cdSRC \cite{cdSRC_compare}; 5) LRR \cite{liu2010robust}; 6) LGIDL \cite{he2016low}.

The above competitors can roughly be divided into three categories: SVM-based, SR-based, and LRR-based methods. To be specific, the classic Support Vector Machine (SVM) is a great classifier which has been widely applied in HSI classification. And another powerful SVM-based method, SVM with composite kernel (SVMCK), has achieved promising classification accuracy due to incorporating the contextual information into the kernels. Furthermore, we also take two SR-based classification algorithms into account. The first one is joint robust sparse representation classifier (JRSRC), which makes these pixels in neighboring regions represented jointly by some common training samples with the same sparse coefficients. An advantage for JRSRC is that it is robust to the HSI outliers. The second is class-dependent sparse representation classifier (cdSRC) , which effectively integrates the idea of KNN into SRC in a class-wise manner and characterizes both Euclidean distance and correlation information between training and testing set. Finally, these LRR-based approaches are the original LRR and LGIDL. Among them, the LGIDL employs superpixel segmentation to obtain the adaptive spatial correlation regions and yields fairly competitive performance.

\textbf{2). Parameter Settings}: Every method is repeated ten times to avoid the bias due to the random sampling. All free parameters of these algorithms are determined via cross validation, using training data only. For SVM-based comparison methods, we choose RBF $K(x_i,x_j)=exp(-\gamma \lVert x_i-x_j\rVert^2)$ as the kernel function of SVM, and the optimal parameters $C$ and $\gamma$ are tuned by grid search algorithm. The one vs. one strategy is applied in the implementation of SVM. Specifically, the parameters of SVM are $C=2000$, $\gamma=0.1$ for Indian Pines, $C=1500$, $\gamma=0.08$ for Pavia University, and $C=4000$, $\gamma=0.001$ for Salinas. For SVMCK, we select the mean spectral values of square patches as the spatial feature, and employ the weighted summation kernel to balance the spatial and spectral components. The patch size $T$ and kernel weight $\mu$ for three datasets are $\{T=15,\ \mu=0.7\}$, $\{T=5,\ \mu=0.8\}$, and $\{T=50,\ \mu=0.4\}$ . Moreover, the optimal parameter settings of JRSRC and LGIDL are followed as \cite{JRSRC_compare} and \cite{he2016low}, respectively. For the proposed LSLRR, the corresponding parameters are set as $\{\lambda=20,\ \alpha=0.8,\ \beta=0.6,\ m=25\}$, $\{\lambda=10,\ \alpha=0.3,\ \beta=1.2,\ m=15\}$, $\{\lambda=10,\ \alpha=1,\ \beta=0.4,\ m=40\}$ for three HSI datasets, respectively

\textbf{3). Evaluation indexes}: We adopt three quantitative metric, overall accuracy (OA), average accuracy (AA) and kappa coefficient ($\kappa$), to evaluate the performance of different classification methods. Specifically, OA index denotes the percentage of HSI pixels which are classified correctly. AA index refers to the average value of accuracy of each class.
However, both OA and AA index only involve the errors of commission and they do not cover the user accuracy. The kappa coefficient ($\kappa$), a more reasonable measurement, not only involves the errors of commission but also the errors of omission.

\subsection{Experimental Results and Analyses}
\textbf{Indian Pines}: We randomly select 10\% labeled samples in each class as the training set, and the rest as the testing set. Table \ref{tab:inp} demonstrates the final classification performance (i.e., the accuracy for each category, OA, AA and kappa coefficient $\kappa$) for the Indian Pines dataset. The corresponding classification maps of each algorithm are shown in Fig. \ref{fig:inp_maps}. Among these comparison algorithms of HSI classification, SVM and LRR are pixel-wise classification methods which only utilize the spectral feature. Other algorithms (SVMCK, JRSRC, LGIDL and LSLRR) combine both spectral and spatial information to classify HSI data. One can be seen easily from Table \ref{tab:inp} that classification accuracy of SVM and LRR is far lower (OA decreases at least 10\%) than that of the other methods. This indicates that the contextual feature can bring a great help for HSI classification. In addition, SVM outperforms the LRR a lot, which verifies the popular SVM is a superior classification algorithm. For classification accuracy of every class in Table \ref{tab:inp}, LGIDL achieves the best result for the 2-th class. JRSRC achieves the best result for the 3-th class. cdSRC achieves the best result for the 6-th class. The proposed LSLRR also obtains the highest accuracy in most classes. Furthermore, the classification OA of the proposed LSLRR improves more than 20\% compared with the classical LRR. This is because LCC helps LRR to capture the local feature and SPS makes the solution $\hat{Z}$ close to ideal block-diagonal matrix. Moreover, Table \ref{tab:inp} also obviously demonstrates that LSLRR has achieved the best performance than all other comparison methods. Fig. \ref{fig:percent} (a) illustrates the classification accuracy of various methods when different number of samples are considered as training set. It can be clearly observed that the classification performance of SVM and LRR is the worst. And other classification methods all have a promising performance. Among these, the proposed LSLRR yields the best classification results.

\textbf{Pavia University}: 5\% of labeled HSI pixels are chose to be training set, and the remaining 95\% is used for testing. In order to compare the experimental results quantitatively and visually, Table \ref{tab:pav} and Fig. \ref{fig:pav_maps} exhibit the classification performance of Pavia University, and the corresponding visual maps of all methods, respectively. As is shown in Table \ref{tab:pav} and Fig. \ref{fig:pav_maps}, only a small number of HSI pixels are classified wrongly, and the classification accuracy of LSLRR is the highest in three evaluation indexes. Except for the 9-th class, LSLRR achieves the best results for other 8 classes. This indicates that LSLRR is an effective and superior approach to classify HSIs. After incorporating the spatial characteristics into the composite kernels, SVMCK yields better classsification results in almost all classes compared with SVM. Similarly, the OA of original LRR is the lowest, and the main samples which is wrongly classified is class 3, 4, 5, and 6. As is seen from Fig. \ref{fig:pav_maps} (f), there are so many red pixels (class 2) in the blue regions (class 6). Through improving LRR by two powerful techniques, LSLRR achieves the OA of 99.9\% in the 6-th class. Compared with LRR, OA of LSLRR improves nearly 12\%, and kappa coefficient ($\kappa$) of LSLRR improves more than 16\%. Furthermore, we also investigate the influence of different number of training pixels on classification accuracy for Pavia University set. And the corresponding figure is demonstrated in Fig. \ref{fig:percent} (b). Interestingly, the curve of LSLRR is the highest while that of LRR is the lowest, which reveals the improvement of LSLRR for LRR is successful.

\textbf{Salinas}: Similar to Pavia University, 5\% pixels are selected to train classification model and the rest 95\% is as the testing set. The classification accuracy of comparing methods and LSLRR are displayed in Table \ref{tab:sal}. For the purpose of visualization, the classification maps are illustrated in Fig. \ref{fig:sal_maps}. From the visual maps, the most classified-wrongly pixels are in the dark-blue (class 8) and dark-green (class 15) regions. This is because the land surfaces of the 8-th and the 15-th classes have homologous properties, and the corresponding spectral reflectance curves are very similar. In addition, it is easy to observe that the proposed LSLRR yields the best accuracy compared with other methods, which justifys the effectiveness of LSLRR. From Table \ref{tab:sal}, we can see that the classification accuracy of most classes is more than 99\% and all OA is not lower than 94\%. Moreover, Fig. \ref{fig:percent} (c) exhibits the overall accuracy of different methods for Salinas scene versus the percentage of training samples. This clearly displays that LSLRR can still obtain the best performance although a small number of pixels are used for training set.

Fig. \ref{fig:para_m} exhibits the OA of three HSI datasets when the value of parameter $m$ changes. Other crucial parameters are followed as subsection \ref{ssec:Experimental Setups}. $m$ is an important parameter to control the weight of spatial information in the LCC. From Fig. \ref{fig:para_m}, we can get that the optimal value of $m$ is 25, 15, and 40 for Indian Pines, Pavia University, and Salinas, respectively. The way we employ to measure the spatial similarity is by  Eucliden distance, which is more suitable to pixels of the same class distributing in a square or circular shape. As is seen from Fig. \ref{fig:para_m}, the shapes of many classes in Pavia Unversity are slender, while Salinas has many pixels whose distribution is more uniform. Therefore, the most appropriate $m$ for Salinas is the largest, and that for Pavia Unversity is the smallest. In summary, the large value of $m$ is more reasonable for the HSI dataset, which has higher compactness for each class.

Fig. \ref{fig:para_alpha} and Fig. \ref{fig:para_beta} illustrate the overall accuracy of three HSI datasets under different values of parameter $\alpha$ and $\beta$, respectively. When investigating the influence of classification accuracy about parameter $\alpha$ or $\beta$, other parameters are set as the optimal values. Obviously, the optimal values of $\alpha$ for three datasets are 0.8, 0.3 and 1.0, respectively. When the locality constraint criterion (LCC) is not added, i.e. $\alpha=0$, the classification accuracy decreases a lot comparing with the highest OA for all three datasets. Especially for Indian Pines, OA decreases more than 12\%. This indicates LCC is extremely important for the proposed LSLRR. Furthermore, one can be easily seen that the optimal values of $\beta$ for three datasets are 0.6, 1.2 and 0.4, respectively. Similarly, when $\beta=0$, classification accuracy is very low. And the OA index improves so fast when the value of $\beta$ starts to increase from 0. It demonstrates the importance of structure preserving strategy (SPS). To sum up, both LCC and SPS can provide a great deal of help to improve significantly the classification accuracy.

\subsection{Comparison of Running Time}
\label{ssec:running time}
As follows, in order to testify the efficiency of the proposed LSLRR, we use running time to compare the computational complexity of all algorithms. Indian Pines dataset is considered as an example, and 10\% of labeled pixels of each class are used for training model. The experiments are conducted in MATLAB R2015a on a PC of Intel Core i7-3770 3.40GHz CPU with 32 GB RAM. TABLE \ref{table:running time} shows OA, AA, kappa coefficient and running time of every methods. According to it, the time consuming of SVM and SVMCK are the least, but their classification accuracy is not high enough comparing with JRSRC, cdSRC, LGIDL and LSLRR. JRSRC and LGIDL can obtain promising classification performance, but the running time is too long. For the proposed LSLRR, it is computationally acceptable and the classification accuracy is the highest.

\begin{table}[htbp] \small
    \centering
        \caption{Running time of different HSI classification methods}
            \begin{tabular}{p{0.5in}<{\centering}  p{0.5in}<{\centering}  p{0.5in}<{\centering}  p{0.5in}<{\centering}  p{0.5in}<{\centering} }
                \toprule
                Methods & OA(\%)& AA(\%)& Kappa  & Time(s)\\ \midrule
                SVM     & 81.67 & 78.60 & 0.7902 &   4.23\\
                SVMCK   & 91.93 & 86.76 & 0.9076 &   6.17\\
                JRSRC   & 92.36 & 83.99 & 0.9124 & 328.62\\
                cdSRC   & 93.61 & 90.32 & 0.9270 & 118.86\\
                LRR     & 70.47 & 54.19 & 0.6545 & 242.37\\
                LGIDL   & 94.52 & 87.60 & 0.9374 & 382.13\\
                LSLRR   & 95.63 & 92.74 & 0.9512 & 336.25\\
                \bottomrule
            \end{tabular}
    \label{table:running time}
\end{table}

\section{Conclusion}
\label{sec:conclusion}
In this paper, a novel locality and structure regularized low rank representation (LSLRR) is proposed to classify hyperspectral images. In order to overcome the drawbacks of traditional low rank representation (LRR), LSLRR introduces two key techniques, locality constraint criterion (LCC) and structure preserving strategy (SPS), to improve LRR and make it more suitable for HSI classification. In LSLRR, a new similarity metric combining both spatial and spectral characteristics is first presented. And then LCC utilizes the new similarity metric to make HSI pixels with large distance have a small similarity, which can easily capture the local structure. Besides, SPS makes the solution of LSLRR close to a class-wise block-diagonal matrix. Finally, the classification results can be easily obtained without any complex classifiers. Extensive experiments on three public HSI datasets are carried out to evaluate the performance of the proposed LSLRR. And the experimental results show that LSLRR outperforms other state-of-the-art comparison methods.

\small
\bibliographystyle{IEEEtran}
\bibliography{strings1}

\begin{thebibliography}{10}
\providecommand{\url}[1]{#1}
\csname url@samestyle\endcsname
\providecommand{\newblock}{\relax}
\providecommand{\bibinfo}[2]{#2}
\providecommand{\BIBentrySTDinterwordspacing}{\spaceskip=0pt\relax}
\providecommand{\BIBentryALTinterwordstretchfactor}{4}
\providecommand{\BIBentryALTinterwordspacing}{\spaceskip=\fontdimen2\font plus
\BIBentryALTinterwordstretchfactor\fontdimen3\font minus
  \fontdimen4\font\relax}
\providecommand{\BIBforeignlanguage}[2]{{%
\expandafter\ifx\csname l@#1\endcsname\relax
\typeout{** WARNING: IEEEtran.bst: No hyphenation pattern has been}%
\typeout{** loaded for the language `#1'. Using the pattern for}%
\typeout{** the default language instead.}%
\else
\language=\csname l@#1\endcsname
\fi
#2}}
\providecommand{\BIBdecl}{\relax}
\BIBdecl

\bibitem{wang2017locality}
Q.~Wang, Z.~Meng, and X.~Li, ``Locality adaptive discriminant analysis for
  spectral--spatial classification of hyperspectral images,'' \emph{IEEE
  Geosci. Remote Sens. Lett.}, vol.~14, no.~11, pp. 2077--2081, 2017.

\bibitem{hughes1968mean}
G.~Hughes, ``On the mean accuracy of statistical pattern recognizers,''
  \emph{IEEE Trans. Inf. Theory}, vol.~14, no.~1, pp. 55--63, 1968.

\bibitem{SVMCK_compare}
G.~Camps-Valls, L.~Gomez-Chova, J.~Mu{\~n}oz-Mar{\'\i}, J.~Vila-Franc{\'e}s,
  and J.~Calpe-Maravilla, ``Composite kernels for hyperspectral image
  classification,'' \emph{IEEE Geosci. Remote Sens. Lett.}, vol.~3, no.~1, pp.
  93--97, 2006.

\bibitem{camps2010spatio}
G.~Camps-Valls, N.~Shervashidze, and K.~M. Borgwardt, ``Spatio-spectral remote
  sensing image classification with graph kernels,'' \emph{IEEE Geosci. Remote
  Sens. Lett.}, vol.~7, no.~4, pp. 741--745, 2010.

\bibitem{liu2010robust}
G.~Liu, Z.~Lin, and Y.~Yu, ``Robust subspace segmentation by low-rank
  representation,'' in \emph{Proceedings of the 27th international conference
  on machine learning (ICML-10)}, 2010, pp. 663--670.

\bibitem{li2014learning}
Y.~Li, J.~Liu, Z.~Li, Y.~Zhang, H.~Lu, S.~Ma \emph{et~al.}, ``Learning low-rank
  representations with classwise block-diagonal structure for robust face
  recognition.'' in \emph{AAAI}, 2014, pp. 2810--2816.

\bibitem{LRR_image}
L.~Li, S.~Li, and Y.~Fu, ``Learning low-rank and discriminative dictionary for
  image classification,'' \emph{Image Vision Comput.}, vol.~32, no.~10, pp.
  814--823, 2014.

\bibitem{liu2013robust}
G.~Liu, Z.~Lin, S.~Yan, J.~Sun, Y.~Yu, and Y.~Ma, ``Robust recovery of subspace
  structures by low-rank representation,'' \emph{IEEE Trans. Pattern Anal.
  Mach. Intell.}, vol.~35, no.~1, pp. 171--184, 2013.

\bibitem{zhou2013moving}
X.~Zhou, C.~Yang, and W.~Yu, ``Moving object detection by detecting contiguous
  outliers in the low-rank representation,'' \emph{IEEE Trans. Pattern Anal.
  Mach. Intell.}, vol.~35, no.~3, pp. 597--610, 2013.

\bibitem{wang2018getnet}
Q.~Wang, Z.~Yuan, and X.~Li, ``Getnet: A general end-to-end two-dimensional cnn
  framework for hyperspectral image change detection,'' \emph{IEEE Trans.
  Geosci. Remote Sens.}, 2018.

\bibitem{sun2014structured}
X.~Sun, Q.~Qu, N.~M. Nasrabadi, and T.~D. Tran, ``Structured priors for
  sparse-representation-based hyperspectral image classification,'' \emph{IEEE
  Geosci. Remote Sens. Lett.}, vol.~11, no.~7, pp. 1235--1239, 2014.

\bibitem{mei2016spectral}
S.~Mei, Q.~Bi, J.~Ji, J.~Hou, and Q.~Du, ``Spectral variation alleviation by
  low-rank matrix approximation for hyperspectral image analysis,'' \emph{IEEE
  Geosci. Remote Sens. Lett.}, vol.~13, no.~6, pp. 796--800, 2016.

\bibitem{wang2016salient}
Q.~Wang, J.~Lin, and Y.~Yuan, ``Salient band selection for hyperspectral image
  classification via manifold ranking,'' \emph{IEEE Trans. Neural Netw. Learn.
  Syst.}, vol.~27, no.~6, pp. 1279--1289, 2016.

\bibitem{chen2011hyperspectral}
Y.~Chen, N.~M. Nasrabadi, and T.~D. Tran, ``Hyperspectral image classification
  using dictionary-based sparse representation,'' \emph{IEEE Trans. Geosci.
  Remote Sens.}, vol.~49, no.~10, pp. 3973--3985, 2011.

\bibitem{srinivas2013exploiting}
U.~Srinivas, Y.~Chen, V.~Monga, N.~M. Nasrabadi, and T.~D. Tran, ``Exploiting
  sparsity in hyperspectral image classification via graphical models,''
  \emph{IEEE Geosci. Remote Sens. Lett.}, vol.~10, no.~3, pp. 505--509, 2013.

\bibitem{zhang2014nonlocal}
H.~Zhang, J.~Li, Y.~Huang, and L.~Zhang, ``A nonlocal weighted joint sparse
  representation classification method for hyperspectral imagery,'' \emph{IEEE
  J. Sel. Topics Appl. Earth Observ. Remote Sens.}, vol.~7, no.~6, pp.
  2056--2065, 2014.

\bibitem{li2016hyperspectral}
C.~Li, Y.~Ma, X.~Mei, C.~Liu, and J.~Ma, ``Hyperspectral image classification
  with robust sparse representation,'' \emph{IEEE Geosci. Remote Sens. Lett.},
  vol.~13, no.~5, pp. 641--645, 2016.

\bibitem{li2015efficient}
J.~Li, H.~Zhang, and L.~Zhang, ``Efficient superpixel-level multitask joint
  sparse representation for hyperspectral image classification,'' \emph{IEEE
  Trans. Geosci. Remote Sens.}, vol.~53, no.~10, pp. 5338--5351, 2015.

\bibitem{fu2017adaptive}
W.~Fu, S.~Li, L.~Fang, and J.~A. Benediktsson, ``Adaptive spectral--spatial
  compression of hyperspectral image with sparse representation,'' \emph{IEEE
  Trans. Geosci. Remote Sens.}, vol.~55, no.~2, pp. 671--682, 2017.

\bibitem{gan2018multiple}
L.~Gan, J.~Xia, P.~Du, and J.~Chanussot, ``Multiple feature kernel sparse
  representation classifier for hyperspectral imagery,'' \emph{IEEE Trans.
  Geosci. Remote Sens.}, 2018.

\bibitem{fang2014spectral}
L.~Fang, S.~Li, X.~Kang, and J.~A. Benediktsson, ``Spectral--spatial
  hyperspectral image classification via multiscale adaptive sparse
  representation,'' \emph{IEEE Trans. Geosci. Remote Sens.}, vol.~52, no.~12,
  pp. 7738--7749, 2014.

\bibitem{tang2016manifold}
Y.~Y. Tang and H.~Yuan, ``Manifold-based sparse representation for
  hyperspectral image classification,'' in \emph{Handbook of Pattern
  Recognition and Computer Vision}.\hskip 1em plus 0.5em minus 0.4em\relax
  World Scientific, 2016, pp. 331--350.

\bibitem{li2016survey}
W.~Li and Q.~Du, ``A survey on representation-based classification and
  detection in hyperspectral remote sensing imagery,'' \emph{Pattern Recognit.
  Lett.}, vol.~83, pp. 115--123, 2016.

\bibitem{wang2018optimal}
Q.~Wang, F.~Zhang, and X.~Li, ``Optimal clustering framework for hyperspectral
  band selection,'' \emph{IEEE Trans. Geosci. Remote Sens.}, 2018.

\bibitem{qu2014abundance}
Q.~Qu, N.~M. Nasrabadi, and T.~D. Tran, ``Abundance estimation for bilinear
  mixture models via joint sparse and low-rank representation,'' \emph{IEEE
  Trans. Geosci. Remote Sens.}, vol.~52, no.~7, pp. 4404--4423, 2014.

\bibitem{zhao2015hyperspectral}
Y.-Q. Zhao and J.~Yang, ``Hyperspectral image denoising via sparse
  representation and low-rank constraint,'' \emph{IEEE Trans. Geosci. Remote
  Sens.}, vol.~53, no.~1, pp. 296--308, 2015.

\bibitem{shi2015domain}
Q.~Shi, B.~Du, and L.~Zhang, ``Domain adaptation for remote sensing image
  classification: A low-rank reconstruction and instance weighting label
  propagation inspired algorithm,'' \emph{IEEE Trans. Geosci. Remote Sens.},
  vol.~53, no.~10, pp. 5677--5689, 2015.

\bibitem{yuan2015low}
Y.~Yuan, M.~Fu, and X.~Lu, ``Low-rank representation for 3d hyperspectral
  images analysis from map perspective,'' \emph{Signal Process.}, vol. 112, pp.
  27--33, 2015.

\bibitem{sumarsono2016low}
A.~Sumarsono and Q.~Du, ``Low-rank subspace representation for supervised and
  unsupervised classification of hyperspectral imagery,'' \emph{IEEE J. Sel.
  Topics Appl. Earth Observ. Remote Sens.}, vol.~9, no.~9, pp. 4188--4195,
  2016.

\bibitem{soltani2015spatial}
A.~Soltani-Farani, H.~R. Rabiee, and S.~A. Hosseini, ``Spatial-aware dictionary
  learning for hyperspectral image classification,'' \emph{IEEE Trans. Geosci.
  Remote Sens.}, vol.~53, no.~1, pp. 527--541, 2015.

\bibitem{he2016low}
Z.~He, L.~Liu, R.~Deng, and Y.~Shen, ``Low-rank group inspired dictionary
  learning for hyperspectral image classification,'' \emph{Signal Process.},
  vol. 120, pp. 209--221, 2016.

\bibitem{jia2015spectral}
S.~Jia, X.~Zhang, and Q.~Li, ``Spectral--spatial hyperspectral image
  classification using $\ell_{1/2}$ regularized low-rank representation and
  sparse representation-based graph cuts,'' \emph{IEEE J. Sel. Topics Appl.
  Earth Observ. Remote Sens.}, vol.~8, no.~6, pp. 2473--2484, 2015.

\bibitem{rubinstein2010dictionaries}
R.~Rubinstein, A.~M. Bruckstein, and M.~Elad, ``Dictionaries for sparse
  representation modeling,'' \emph{Proc. IEEE}, vol.~98, no.~6, pp. 1045--1057,
  2010.

\bibitem{engan1999method}
K.~Engan, S.~O. Aase, and J.~H. Husoy, ``Method of optimal directions for frame
  design,'' in \emph{Acoustics, Speech, and Signal Processing, 1999.
  Proceedings., 1999 IEEE International Conference on}, vol.~5.\hskip 1em plus
  0.5em minus 0.4em\relax IEEE, 1999, pp. 2443--2446.

\bibitem{lesage2005learning}
S.~Lesage, R.~Gribonval, F.~Bimbot, and L.~Benaroya, ``Learning unions of
  orthonormal bases with thresholded singular value decomposition,'' in
  \emph{Acoustics, Speech, and Signal Processing, 2005.
  Proceedings.(ICASSP'05). IEEE International Conference on}, vol.~5.\hskip 1em
  plus 0.5em minus 0.4em\relax IEEE, 2005, pp. v--293.

\bibitem{vidal2005generalized}
R.~Vidal, Y.~Ma, and S.~Sastry, ``Generalized principal component analysis
  (gpca),'' \emph{IEEE Trans. Pattern Anal. Mach. Intell.}, vol.~27, no.~12,
  pp. 1945--1959, 2005.

\bibitem{aharon2006rm}
M.~Aharon, M.~Elad, and A.~Bruckstein, ``$ rm k $-svd: An algorithm for
  designing overcomplete dictionaries for sparse representation,'' \emph{IEEE
  Trans. Signal Process.}, vol.~54, no.~11, pp. 4311--4322, 2006.

\bibitem{wang2014spatial}
Z.~Wang, N.~M. Nasrabadi, and T.~S. Huang, ``Spatial--spectral classification
  of hyperspectral images using discriminative dictionary designed by learning
  vector quantization,'' \emph{IEEE Trans. Geosci. Remote Sens.}, vol.~52,
  no.~8, pp. 4808--4822, 2014.

\bibitem{landgrebe2002hyperspectral}
D.~Landgrebe, ``Hyperspectral image data analysis,'' \emph{IEEE Signal Process.
  Mag.}, vol.~19, no.~1, pp. 17--28, 2002.

\bibitem{chakrabarti2011statistics}
A.~Chakrabarti and T.~Zickler, ``Statistics of real-world hyperspectral
  images,'' in \emph{Computer Vision and Pattern Recognition (CVPR), 2011 IEEE
  Conference on}.\hskip 1em plus 0.5em minus 0.4em\relax IEEE, 2011, pp.
  193--200.

\bibitem{lin2011linearized}
Z.~Lin, R.~Liu, and Z.~Su, ``Linearized alternating direction method with
  adaptive penalty for low-rank representation,'' in \emph{Advances in neural
  information processing systems}, 2011, pp. 612--620.

\bibitem{lin2010augmented}
Z.~Lin, M.~Chen, and Y.~Ma, ``The augmented lagrange multiplier method for
  exact recovery of corrupted low-rank matrices,'' \emph{arXiv preprint
  arXiv:1009.5055}, 2010.

\bibitem{tang2014structure}
K.~Tang, R.~Liu, Z.~Su, and J.~Zhang, ``Structure-constrained low-rank
  representation,'' \emph{IEEE Trans. Neural Netw. Learn. Syst.}, vol.~25,
  no.~12, pp. 2167--2179, 2014.

\bibitem{hyper_dataset}
``Hyperspectral remote sensing scenes,'' Accessed: Jun. 1, 2018.
  [Online].Available: \url{http://www.ehu.eus/ccwintco/index.php/Hyperspectral_
  Remote_Sensing_Scenes}.

\bibitem{SVM_compare}
Y.~Xiao, H.~Wang, and W.~Xu, ``Parameter selection of gaussian kernel for
  one-class svm,'' \emph{IEEE Trans. Cybern.}, vol.~45, no.~5, pp. 941--953,
  2015.

\bibitem{JRSRC_compare}
C.~Li, Y.~Ma, X.~Mei, C.~Liu, and J.~Ma, ``Hyperspectral image classification
  with robust sparse representation,'' \emph{IEEE Geosci. Remote Sens. Lett.},
  vol.~13, no.~5, pp. 641--645, 2016.

\bibitem{cdSRC_compare}
M.~Cui and S.~Prasad, ``Class-dependent sparse representation classifier for
  robust hyperspectral image classification,'' \emph{IEEE Trans. Geosci. Remote
  Sens.}, vol.~53, no.~5, pp. 2683--2695, 2015.

\end{thebibliography}

\begin{IEEEbiography}[{\includegraphics[width=1in,height=1.25in,clip,keepaspectratio]{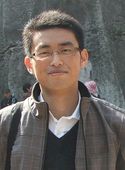}}]{Qi Wang}
(M'15-SM'15) received the B.E. degree in automation and the Ph.D. degree in pattern recognition and intelligent systems from the University of Science and Technology of China, Hefei, China, in 2005  and 2010, respectively.  He is currently a Professor with the School of Computer Science, with the Unmanned System Research Institute, and with the Center for OPTical IMagery Analysis and Learning, Northwestern Polytechnical University, Xi'an, China. His research interests include computer vision and pattern recognition.
\end{IEEEbiography}

\begin{IEEEbiography}[{\includegraphics[width=1in,height=1.25in,clip,keepaspectratio]{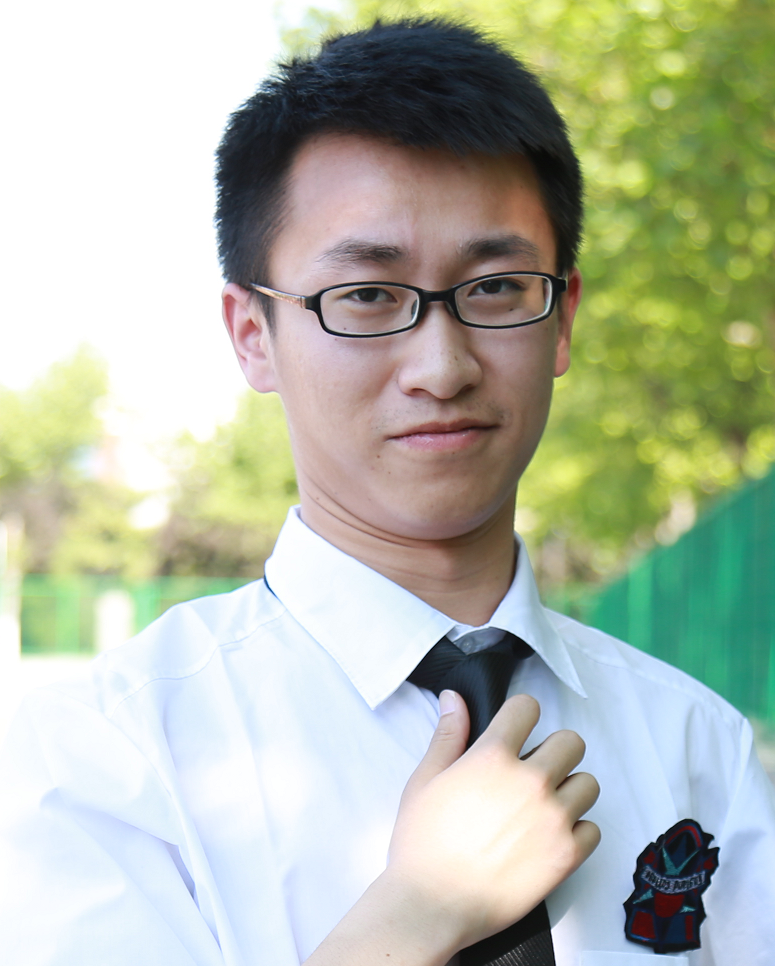}}]{Xiang He}
received the B.E. degree in automation from Northwestern Polytechnical University, Xi'an, China, in 2017. He is currently working toward the M.S. degree in computer science in the Center for OPTical IMagery Analysis and Learning (OPTIMAL), Northwestern Polytechnical University, Xi'an, China.
His research interests include hyperspectral image processing and computer vision.
\end{IEEEbiography}

\begin{IEEEbiographynophoto}{Xuelong Li}
(M'02-SM'07-F'12) is a full professor with the Xi'an Institute of Optics and Precision Mechanics, Chinese Academy of Sciences, Xi'an 710119, Shaanxi, P. R. China and with the University of Chinese Academy of Sciences, Beijing 100049, P. R. China.
\end{IEEEbiographynophoto}
%
%
%




\end{document}